\documentclass[11pt]{article}
\usepackage[final]{acl}
\usepackage{times}
\usepackage{latexsym}
\usepackage[T1]{fontenc}
\usepackage[utf8]{inputenc}
\usepackage{microtype}
\usepackage{inconsolata}
\usepackage{amsmath,amssymb,amsfonts}
\usepackage{adjustbox}
\usepackage{algorithmic}
\usepackage{graphicx}
\usepackage{textcomp}
\usepackage{pgf-pie}
\usepackage{xcolor}
\usepackage{booktabs}
\usepackage{tabularx}
\usepackage{multirow}
\usepackage{hyperref}
\usepackage{listings}
\usepackage{stfloats}
\usepackage{tcolorbox}
\usepackage{tikz}
\usetikzlibrary{calc, arrows.meta, positioning, shadows}
\tcbuselibrary{listingsutf8, breakable}
\usepackage{fancyhdr}
\fancypagestyle{emnlpnumbered}{%
  \fancyhf{}%
  \fancyfoot[C]{\normalfont\thepage}%
}
\fancypagestyle{emnlpfirst}{%
  \fancyhf{}%
  \fancyfoot[C]{%
    \begin{minipage}{\textwidth}
      \centering\normalfont
      \thepage\par\vspace{5pt}%
      {\fontsize{9}{11}\selectfont
      {\itshape Proceedings of the 2026 Conference on Empirical Methods in Natural Language Processing,}\nobreakspace pages 40966--40997\par
      October 24-29, 2026 \textcopyright{}2026 Association for Computational Linguistics\par}%
    \end{minipage}%
  }%
}
\title{PubMedCausal: A Span-Level Annotated Corpus for Causal Relation Extraction in Biomedical Text}

\author{
Ifeoluwa Kunle-John\thanks{Equal contribution. Co-first authors.}\textsuperscript{1,2} \quad 
Josiah Paul\footnotemark[1]\textsuperscript{1,2} \quad  
Oluwatosin Agbaakin\thanks{Equal contribution. Co-second authors.}\textsuperscript{3} \\
\textbf{Peter Aina}\footnotemark[2]\textsuperscript{3} \quad 
\textbf{Ikenna Odezuligbo}\footnotemark[2]\textsuperscript{4} \quad 
\textbf{Sydney Anuyah}\thanks{Project lead.}\textsuperscript{1,3} \\
Edyah Limited\textsuperscript{1}, University of Ibadan, NG\textsuperscript{2} \\
Indiana University, Indianapolis, IN\textsuperscript{3}, Creighton University, Omaha NE\textsuperscript{4} \\
{\small\ttfamily \{ikunlejohn, pjosiah, sanuyah\}@edyahlimited.com.ng, \quad \{ikunle-john252399, pjosiah539\}@stu.ui.edu.ng} \\
{\small\ttfamily oluwatosin.agbaakin@alumni.iu.edu, petaina@iu.edu, ikennaodezuligbo@creighton.edu, sanuyah@iu.edu }
}

\definecolor{purpleBox}{RGB}{238,232,255}
\definecolor{cyanBox}{RGB}{225,247,246}
\definecolor{grayBox}{RGB}{244,244,244}
\definecolor{blueBox}{RGB}{226,241,254}
\definecolor{greenBox}{RGB}{225,247,231}
\definecolor{yellowBox}{RGB}{255,242,207}
\definecolor{pinkBox}{RGB}{255,229,232}
\definecolor{peachBox}{RGB}{255,238,224}
\definecolor{creamBox}{RGB}{255,247,221}
\definecolor{deepPurple}{RGB}{79,55,154}
\definecolor{deepTeal}{RGB}{24,130,145}
\definecolor{deepGreen}{RGB}{58,150,85}
\definecolor{deepRed}{RGB}{208,70,82}
\definecolor{deepOrange}{RGB}{238,128,39}
\definecolor{deepGold}{RGB}{231,171,24}
\definecolor{deepGray}{RGB}{83,86,89}
\definecolor{arrowGray}{RGB}{135,134,145}

\begin{document}

\maketitle
\setcounter{page}{40966}
\pagestyle{emnlpnumbered}
\thispagestyle{emnlpfirst}

\begin{abstract}
Causal relation extraction (CRE) is central to biomedical text mining, but current resources often conflate causal relations with broader associations, restrict annotation to sentence-level examples, or focus mainly on explicit causal cues. This limits their usefulness for evaluating whether models can recover causal claims as they are actually expressed in biomedical text. We introduce \textsc{PubMedCausal}, a span-level annotated corpus for biomedical CRE built from PubMed abstracts. The corpus contains 30,000 paragraph-level rows, including 3,945 causal rows and 6,491 adjudicated cause--effect pairs. Each causal relation is annotated with full-text cause and effect spans, causality type, and sententiality, enabling evaluation of both causal detection and full-span causal extraction. We benchmark discriminative encoders and open-source generative models across detection and extraction settings. For causal detection, biomedical encoders are strongest, with PubMedBERT reaching an F$_1$ score of 0.7391. For span-level extraction, the best generative baseline is DeepSeek-R1-Distill-Qwen-32B with few-shot prompting, reaching a Cosine Pair F$_1$ of 0.6765. We further test transfer learning by evaluating PubMedCausal-trained encoders on external causal relation datasets, showing that the resource supports cross-dataset evaluation. Our results show that biomedical CRE remains difficult under class imbalance, long causal spans, implicit causality, inter-sentential relations, and prompt sensitivity. \href{https://github.com/josiahpaul07/PubMedCausal_Exp}{Code and Data can be found here}
\end{abstract}

\begin{figure}[h]
    \centering
    \includegraphics[width=\columnwidth]{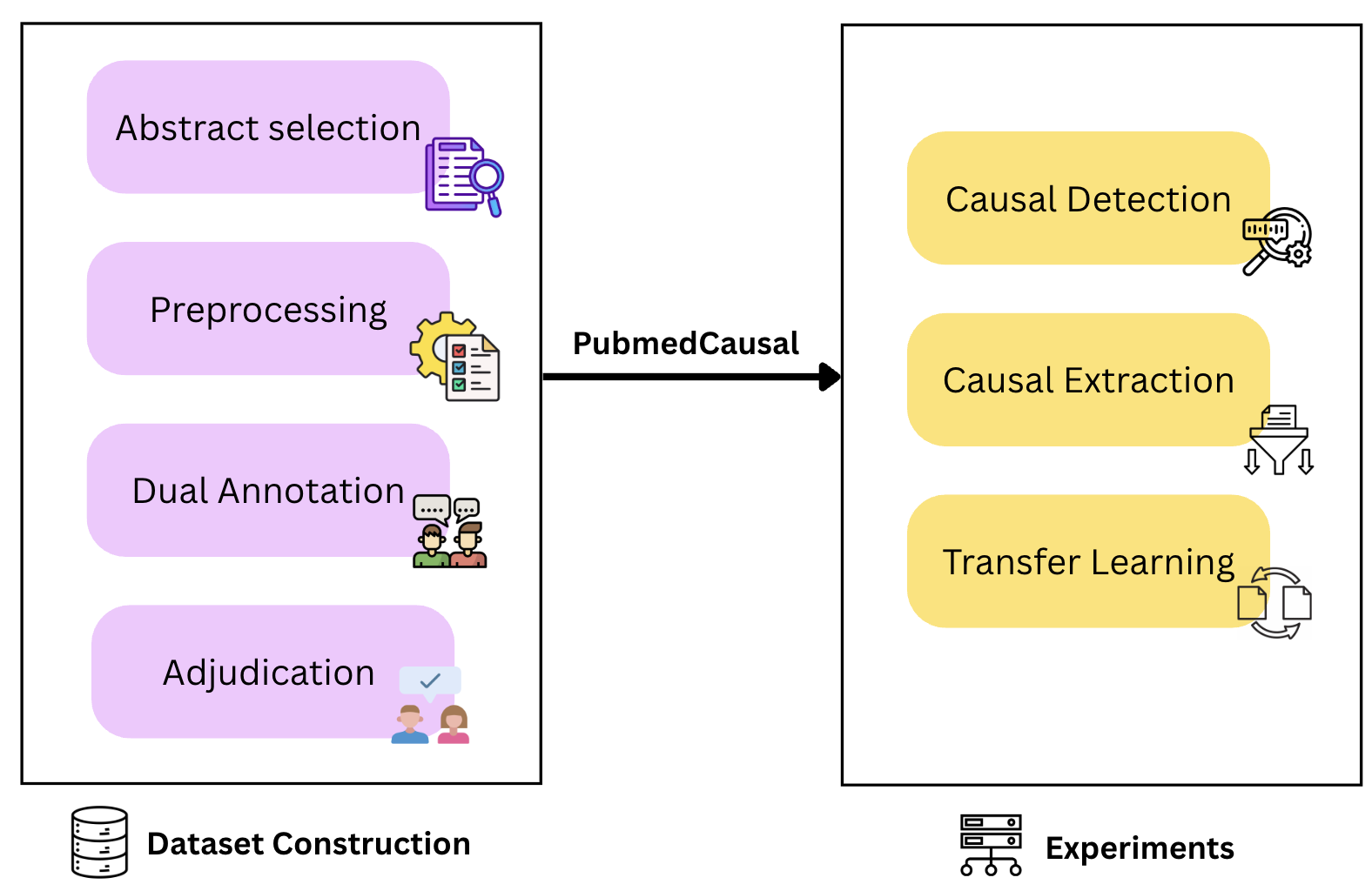}
    \caption{Overview of our work}
    \label{fig:overview}
\end{figure}

\section{Introduction}
\label{sec:intro}
\begin{table*}[t!]
\centering
\tiny
\caption{Coverage matrix. $\checkmark$ explicit; $\sim$ partially supported; $\times$ none.
\textbf{Pair} = structured cause--effect pair annotation;
\textbf{Exp} = explicit causality;
\textbf{Imp} = implicit causality;
\textbf{Intra} = intra-sentential relations;
\textbf{Inter} = inter-sentential relations;
\textbf{Span} = span-level (non-atomic) arguments.}
\label{tab:coverage_matrix}
\setlength{\tabcolsep}{4pt}
\begin{tabularx}{\textwidth}{
    >{\raggedright\arraybackslash}p{2.8cm}
    >{\raggedright\arraybackslash}p{1.8cm}
    >{\raggedright\arraybackslash}p{4.2cm}
    *{6}{>{\centering\arraybackslash}X}}
\toprule
\textbf{Dataset}
  & \textbf{Domain} & \textbf{Size \& Key Features}
  & \textbf{Pair} & \textbf{Exp} & \textbf{Imp}
  & \textbf{Intra} & \textbf{Inter} & \textbf{Span} \\
\midrule
AltLex~\cite{hidey2016identifying}
  & General / News
  & 1,027 sents.; 456 causal / 571 non-causal; causal connective labels
  & $\times$ & $\sim$ & $\sim$
  & $\checkmark$ & $\sim$ & $\times$ \\
\addlinespace[3pt]
SemEval 2010 Task 8~\cite{hendrickx-etal-2010-semeval}
  & General / News
  & 10.7K sents.; 9 relation types; 1,003 cause--effect pairs
  & $\checkmark$ & $\checkmark$ & $\times$
  & $\checkmark$ & $\times$ & $\times$ \\
\addlinespace[3pt]
Causal News Corpus~\cite{tan2022causal}
  & News
  & 3,559 event sents.; binary causal labels; protest-domain sources
  & $\times$ & $\checkmark$ & $\sim$
  & $\checkmark$ & $\sim$ & $\times$ \\
\addlinespace[3pt]
BioCause~\cite{Mihaila2013}
  & Biomedical
  & 851 causal relations; 19 full-text articles; infectious disease subdomain
  & $\checkmark$ & $\checkmark$ & $\times$
  & $\checkmark$ & $\checkmark$ & $\sim$ \\
\addlinespace[3pt]
FinCausal~\cite{mariko2020fincausal}
  & Financial
  & 29K segments; 3,405 docs; 4,029 cause tags; 5,350 quantitative fact tags
  & $\checkmark$ & $\checkmark$ & $\sim$
  & $\checkmark$ & $\times$ & $\checkmark$ \\
\addlinespace[3pt]
Causal TimeBank~\cite{mirza-etal-2014-annotating}
  & General / News
  & 6,811 events; 318 explicit causal links with source$\rightarrow$target annotation
  & $\checkmark$ & $\checkmark$ & $\times$
  & $\checkmark$ & $\checkmark$ & $\times$ \\
\midrule
\textbf{PubMedCausal (Ours)}
  & Biomedical
  & 30K paragraph rows (PubMed); 3,945 causal rows; 6,491 cause--effect pairs
  & $\checkmark$ & $\checkmark$ & $\checkmark$
  & $\checkmark$ & $\checkmark$ & $\checkmark$ \\
\bottomrule
\end{tabularx}
\end{table*}
Large Language Models (LLMs) demonstrate strong general language capabilities, yet they continue to struggle with causal extraction (CE), often conflating statistical associations with genuine causal relationships~\cite{bazgir-etal-2025-beyond}. This limitation is especially pronounced for implicit causal relations and for causal relations that extend across sentence boundaries~\cite{anuyah-etal-2025-benchmarking,zadrozny-2025-challenges}. Identifying causality in text requires attention to lexical choice, syntactic composition, discourse relations, and modality, since each can substantially alter the meaning, direction, and strength of a causal claim.

As these linguistic mechanisms recur across domains, they may support a more transferable basis for modeling causality than approaches tied primarily to domain-specific ontologies or knowledge bases~\cite{feder2022causal,singhal2023large,chen2025benchmarking,lai2023keblm}. We use biomedical literature because it includes many causal relations grounded in experimental and clinical evidence. Reliable CE in biomedical contexts can support downstream applications such as drug discovery, disease management, clinical decision-making, and precision medicine~\cite{prosperi2020causal}. However, the abundance of biomedical data presents a growing scale problem for manual annotation; as PubMed alone indexes more than 5,000 new articles per day and contains over 37.6 million records~\cite{sayers2022database, Sayers2025NCBI}. Beyond the published literature, hospitals generate large volumes of electronic health records (EHRs) that may contain clinically relevant causal information, further motivating the need for robust CE methods.

Current automated systems remain limited in their ability to capture this complexity.  We did not design a new model to resolve this challenge. Instead, we provide a resource for evaluating CE, enabling deeper examination of causality and uncertainty types while offering a baseline for future work.

Large-scale resources such as CauseNet and Webis Medical CauseNet employ rule-based extraction strategies to collect millions of cause-effect pairs \cite{heindorf2020causenet}, but these methods remain constrained by precision and by the rigidity of predefined extraction patterns. Existing supervised approaches similarly face important limitations. Relation extraction (RE) datasets frequently conflate causal and associative relationships, while causality-focused datasets are often domain-narrow, limited in scale, or restricted to sentence-level annotations \cite{CRED2025, moghimifar-etal-2020-domain}. In addition, these datasets provide limited coverage of implicit or inter-sentential causality, in favour of explicit causality. Synthetic datasets derived from structural causal models (SCMs), such as COPA~\cite{roemmele2011choice}, offer valuable insights into implicit causality but may introduce statistical patterns that learning algorithms exploit, limiting generalization to naturally occurring text~\cite{ormaniec2025standardizing}. Consequently, there remains a need for large-scale, human-annotated resources that preserve the linguistic structure of causal text.

To address these limitations, we propose an annotation framework for CE centered on textual structure and causal expression, and introduce its first large-scale instantiation, PubMedCausal.

We investigate the following research questions:

\begin{itemize}
    \item \textbf{RQ1:} Can current models accurately detect causal relations in text when causal instances are substantially outnumbered by non-causal instances?

    \item \textbf{RQ2:} How effectively can current models extract full-span cause-effect relations from real-world biomedical paragraphs?

    \item \textbf{RQ3:} How does fine-tuning on annotations that preserve textual structure affect extraction performance and robustness across prompting settings?

    \item \textbf{RQ4:} How do models perform on explicit versus implicit causality, and intra-sentential versus inter-sentential relations?
\end{itemize}

Our main contributions are as follows:

\begin{itemize}

    \item We present PubMedCausal, a rigorously annotated resource containing 30,000 rows of 2--5 sentences from PubMed abstracts and 6,491 cause-effect pairs across 3,945 causal instances.

    \item We annotate cause and effect as full-text spans, preserving contextual information relevant to causal interpretation.

    \item We explicitly distinguish causal and non-causal rows, supporting both causal detection and causal extraction.

    \item We annotate causal relations by causality type and sententiality, enabling targeted evaluation of implicit and cross-sentence causality.
\end{itemize}

\begin{figure*}[t!]
    \centering
    \includegraphics[width=0.95\textwidth]{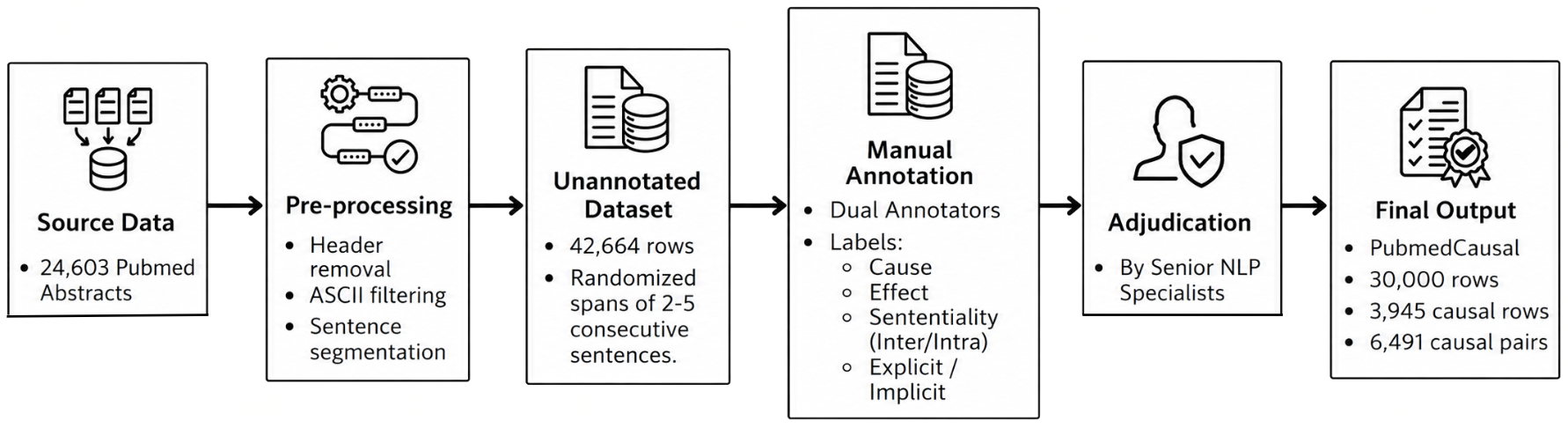}
    \caption{PubMedCausal construction process. Abstracts were retrieved from PubMed, preprocessed, filtered and dual annotated. The final corpus contains 6,491 adjudicated causal relations labeled by sententiality and causality type.}
    \label{fig:PubMedCausal_construction}
\end{figure*}

\section{Related Work}
\label{sec:related}

The motivation of this work is to address the lack of human-annotated resources that distinguish causal relations from broader biomedical associations. Existing biomedical RE datasets such as BC5CDR, GAD, DisGeNET, and BioRED have enabled substantial progress in extracting relations among chemicals, genes, diseases, and other biomedical entities \cite{li2016biocreative,bravo2015extraction,pinero2016disgenet,Luo2022}. However, these resources generally target associative or entity-level relations rather than causal relations, making them insufficient for evaluating whether models can identify causality as expressed in biomedical text.

CE has been studied in general-domain datasets such as AltLex, Causal TimeBank, and the Causal News Corpus \cite{hidey2016identifying,mirza-etal-2014-annotating,tan2022causal}. These resources capture important forms of causal language, but they are either restricted largely to sentence-level relations, or focused on explicit causal cues. Biomedical causal resources are more limited. BioCause annotates causal relations in biomedical articles \cite{Mihaila2013}, MIMICause studies causal relation types in clinical notes \cite{khetan2022mimicause}, and CRED distinguishes causal from non-causal gene--disease relations in PubMed abstracts \cite{CRED2025}. While valuable, these datasets remain limited in scale, domain coverage, or task formulation, particularly for full-span extraction of implicit and cross-sentence causal relations.

Although CauseNet and Webis Medical CauseNet provide broad, automatically extracted causal knowledge, their usefulness for fine-tuning LLMs may be constrained by their relatively predictable patterns, since they were created by rule-based algorithms and they lack non-causal instances ~\cite{heindorf2020causenet, schlatt2022mining}. The presence of only one class makes it less useful for causal detection, and the rule-based patterns can reduce their generalization to real-world contexts.

PubMedCausal addresses these gaps by providing a large-scale, human-annotated biomedical CE resource built from PubMed abstracts. It goes beyond prior biomedical RE datasets by distinguishing causal from non-causal text, and extends existing biomedical CE resources by annotating complete cause and effect spans, covering both explicit and implicit causality, and including intra-sentential as well as inter-sentential relations. Table~\ref{tab:coverage_matrix} summarizes these differences.

\section{Dataset Construction}
\label{sec:methodology}

\begin{table*}[ht]
\centering
\tiny
\caption{Representative annotated examples from PubMedCausal.}
\label{tab:annotation_examples}
\begin{tabular}{>{\raggedright\arraybackslash}p{4.5cm}
                >{\raggedright\arraybackslash}p{3.5cm}
                >{\raggedright\arraybackslash}p{3.5cm}
                >{\centering\arraybackslash}p{1.2cm}
                >{\centering\arraybackslash}p{1.2cm}}
\toprule
\textbf{Span} & \textbf{Cause} & \textbf{Effect} &
\textbf{Sent.} & \textbf{Marker/Causality Type} \\
\midrule
A causal association between media reporting of suicides and subsequent actual suicides has been observed. This study examines whether educating media professionals about responsible reporting can change the quality of reporting. &
media reporting of suicides &
 subsequent actual suicides &
Intra & Explicit \\
\midrule
This study found that sexual victimization significantly increased the likelihood
of a diagnosis of psychosis and therefore suggests that there may be a role for
traumatic experiences in the etiology of psychosis. &
sexual victimization &
a significant increase in the likelihood of a diagnosis of psychosis &
Intra & Implicit \\
\midrule
HyCoSy is a well-tolerated examination and the associated vagal effects are unusual
and generally mild. Consequently, we support its introduction as a first-line procedure
for tubal patency and uterine cavity investigation in infertile women. &
HyCoSy is a well-tolerated examination and the associated vagal effects are unusual
and generally mild &
we support its introduction as a first-line procedure for tubal patency [Pair 1] &
Inter & Implicit \\
\midrule
{[same source as above]} &
HyCoSy is a well-tolerated examination and the associated vagal effects are unusual
and generally mild &
we support its introduction as a first-line procedure for uterine cavity investigation
in infertile women [Pair 2] &
Inter & Implicit \\
\midrule
Sepsis is a common condition encountered in hospital environments. There is no
effective treatment for sepsis, and it remains an important cause of death at
intensive care units. This study aimed to discuss some methods that are available
in clinics, and tests that have been recently developed for the diagnosis of sepsis. &
Sepsis &
death at intensive care units &
Inter & Explicit \\
\bottomrule
\end{tabular}
\end{table*}

\subsection{Source Data}

PubMedCausal was constructed from PubMed abstracts retrieved using the keyword ``causality'' from January 1st to September 3rd, 2025. This query returned 24,603 free full-text abstracts, which were segmented into sentences and then combined into rows with 2--5 successive sentences to support the annotation of both intra-sentential and inter-sentential causal relations. To improve readability and reduce annotation complexity, rows containing numbers or non-ASCII characters were removed, leaving 42,664 candidate rows. Instances without causal relations were retained, since the corpus is intended to support both CE and causal/non-causal classification.

Due to annotation cost and time constraints, we annotated 30,000 rows from this candidate pool. Further details on the sampling strategy, preprocessing decisions, and corpus scope are provided in Appendix \ref{app:sampling_bias}.

\subsection{Annotation Protocol}

\paragraph{Annotators}
We called for applications and offered a test to applicants. The top 10 performers were contacted and trained before commencing the annotation task. All annotators followed the same guideline. The remaining rows were split into batches, and two annotators worked on each batch independently.


\paragraph{Annotation Scheme}
We annotated the text by marking the full cause span and the full effect span for each causal relation. A cause or effect could be a single entity, a noun phrase, or a full clause, depending on how the relation was expressed in the text. If one cause led to more than one separate effect, each cause--effect pair was recorded as a separate row. This applied whether the effects appeared in the same sentence or across different sentences. Where possible, the shared cause span was kept exactly as written in the text, so that separate causal relations were not merged into one broad annotation. Full annotation guidelines and worked examples are provided in Appendix~\ref{app:annotation_guidelines_causality_biomed}.
Each identified relation was annotated along two dimensions:
\begin{enumerate}
    \item \textbf{Sententiality.}
    \textit{Intra-sentential}: both cause and effect spans occur within the same sentence.
    \textit{Inter-sentential}: the spans cross more than one sentence.

    \item \textbf{Causality Type.}
    \textit{Explicit}: the causal relation is directly signaled by a lexical causal marker (see Appendix for the full inventory). 
    \textit{Implicit}: the causal interpretation arises from semantic composition, event structure, or discourse context rather than from a dedicated causal expression.
\end{enumerate}

Table~\ref{tab:annotation_examples} presents representative annotated examples from the corpus, illustrating multi-pair annotation, intra-sentential explicit, intra-sentential implicit, and inter-sentential relations.

\subsection{Adjudication and Inter-Annotator Agreement}

Table~\ref{tab:iaa_metrics} summarises pre-adjudication agreement across the main annotation subtasks. 

To account for boundary variation in extracted spans, we computed token-level
partial match $F_1$. For two spans $a$ and $b$ with token sets $T_a$ and $T_b$,
we define:

\begin{equation}
\small
F_{1,\mathrm{tok}}(a,b)=\frac{2|T_a\cap T_b|}{|T_a|+|T_b|}.
\end{equation}

For texts containing multiple causal relations, annotator outputs were aligned
using bipartite matching. Given two relations
$r_1=(C_1,E_1)$ and $r_2=(C_2,E_2)$, their similarity was:
\begin{equation}
\small
S(r_1,r_2)=\tfrac{1}{2}F_{1,\mathrm{tok}}(C_1,C_2)+\tfrac{1}{2}F_{1,\mathrm{tok}}(E_1,E_2).
\end{equation}

The Hungarian algorithm was then used to find the optimal one-to-one matching
under cost $1-S(r_1,r_2)$. Cause and effect agreement were reported as the mean
token-level $F_1$ over matched relation pairs.

Sententiality classification achieved $\kappa = 0.55$ with raw agreement of $0.98$, again reflecting the effect of label imbalance, since most relations were intra-sentential. Causality expression type achieved $\kappa = 0.61$ with raw agreement of $0.82$, suggesting moderate agreement and confirming the relative difficulty of distinguishing implicit causal relations from explicit ones.

After annotation, an adjudication panel reviewed all disputed cases. They corrected clear errors, adjusted span boundaries when the original span did not fully capture the causal meaning, and reached a final decision on difficult cases.

\begin{table}[t]
\centering
\small
\caption{Pre-adjudication inter-annotator agreement across annotation subtasks.}
\label{tab:iaa_metrics}
\begin{tabular}{lll}
\hline
\textbf{Annotation subtask} & \textbf{Metric} & \textbf{Score} \\
\hline
Causal relation detection & Cohen's $\kappa$ & 0.48 \\
Causal relation detection & Raw agreement & 0.90 \\
Cause span extraction & Partial match $F_1$ & 0.92 \\
Effect span extraction & Partial match $F_1$ & 0.85 \\
Sententiality classification & Cohen's $\kappa$ & 0.55 \\
Sententiality classification & Raw agreement & 0.98 \\
Causality expression type & Cohen's $\kappa$ & 0.61 \\
Causality expression type & Raw agreement & 0.82 \\
\hline
\end{tabular}
\end{table}

\subsection{Corpus Statistics}

As shown in Table~\ref{app:data_details}, 26,055 of the 30,000 processed instances (86.85\%) contain no annotated causal relation. The remaining 3,945 causal instances yielded 6,491 cause--effect relations, with 2,412 instances containing one pair and 1,533 containing multiple pairs.

This imbalance reflects the scope of our causality-query-derived corpus, not causal prevalence in PubMed overall. Even with a causality-oriented query, most processed instances lack annotatable cause--effect relations. Sampling bias and corpus scope are discussed in Appendix \ref{app:bias}.

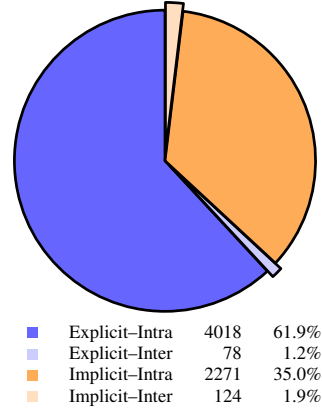
\begin{figure}[t]
\centering
\resizebox{0.55\columnwidth}{!}{%
\begin{tikzpicture}
\pie[
    radius=1.55,
    sum=auto,
    rotate=90,
    hide number,
    hide label,
    explode={0,0.08,0,0.08},
    color={
        blue!60,
        blue!20,
        orange!65,
        orange!25
    }
]{
    4018/Explicit--Intra,
    78/Explicit--Inter,
    2271/Implicit--Intra,
    124/Implicit--Inter
}
\end{tikzpicture}%
}
\vspace{1mm}

{\scriptsize
\begin{tabular}{@{}llrr@{}}
\textcolor{blue!60}{\rule{1.2ex}{1.2ex}} & Explicit--Intra & 4018 & 61.9\% \\
\textcolor{blue!20}{\rule{1.2ex}{1.2ex}} & Explicit--Inter & 78 & 1.2\% \\
\textcolor{orange!65}{\rule{1.2ex}{1.2ex}} & Implicit--Intra & 2271 & 35.0\% \\
\textcolor{orange!25}{\rule{1.2ex}{1.2ex}} & Implicit--Inter & 124 & 1.9\% \\
\end{tabular}
}

\caption{Distribution of causal relations by causality type and sententiality.}
\label{fig:causality_sententiality_pie}
\end{figure}

\section{Experimental Setup}
\label{sec:setup}

\subsection{Model Selection}

We evaluate two model families, encoder-based classifiers and open source LLMs for the causal detection task. The encoder models  used are BERT, BioBERT, PubMedBERT, SciBERT. For the LLMs, we selected several models across different ranges of parameters from 3B to 70B. Larger models (Llama 70B, DeepSeek-R1 32B, DeepSeek 70B, Mixtral 8x7B) were evaluated in zero-shot settings under 4-bit quantization, while smaller models (Llama 3.2-3B, Llama 3.1-8B, Qwen 2.5-7B, Mistral-7B, DeepSeek-V3-3B, DeepSeek-V2.5-7B) were not quantized. In addition, we fine-tuned these smaller LLM classes using LoRA.

On the second task, CE, we evaluated only the LLMs, still holding the same configuration for the first task.

\subsection{Data Splits}

For causal {detection}, the full 30,000-row corpus was split 50:50 into training and test sets using stratified sampling to maintain the empirical class distribution in both partitions. Splitting was performed at the row level. For causal \textit{extraction}, we focused on the 3,945 causal rows and applied a complexity-stratified split. The causal data was first divided into two clusters: single-pair rows (approximately 61\%) and multi-pair rows (approximately 39\%). Each cluster was independently split 50:50; the final training and test sets were formed by merging the corresponding halves. 

\subsection{Prompting Strategies and Evaluation Metrics}

Six prompting strategies were tested for all generative models: zero-shot (ZS), few-shot (FS), chain-of-thought (COT), CoT + FS, ReAct, and least-to-most (L2M) prompting. All prompts are provided in the technical appendix~\ref{app:prompts}.

For causal detection, we report Precision, Recall and F$_1$ on the causal class. For extraction, we use a three-tier evaluation: Exact Match F$_1$ (strict string match), Token Overlap F$_1$ (partial credit for shared tokens), and Cosine Similarity. Cosine Similarity F$_1$ is computed using the embedding model from \citet{deka2022evidence}, fine-tuned on MedNLI and SciNLI, providing domain-appropriate sentence embeddings for biomedical span comparison. 

We apply a pre-specified cosine similarity threshold of 0.75, following recent medical literature mining work \citep{wang2025foundation}. This threshold accommodates minor boundary variation in longer biomedical causal spans while still requiring substantial semantic agreement between predicted and gold spans. 

\section{Baseline Experiments}
\label{sec:results}
We report baseline experiments for the two tasks supported by the corpus. These experiments provide initial points of comparison for future work and illustrate the types of challenges introduced by the dataset, including class imbalance, long causal arguments, implicit causality type, and multi-pair spans. 

For each generative model, all six prompting strategies were evaluated and the best performing configuration per model is reported in this section; complete results are provided in Appendix~\ref{app:full_results}, informing our choice of the best prompt configuration. 

\subsection{Experiment 1: Causal Relation Detection Baselines}

Table~\ref{tab:detection_comparison} reports causal relation detection results on PubMedCausal corpus. This task evaluates whether a model can distinguish paragraphs that contain at least one causal relation from non-causal ones. The encoder models provide the strongest reference baseline, especially PubMedBERT, which obtained the highest F$_1$ score of 0.7391, followed closely by BioBERT with an F$_1$ score of 0.7380. These results suggest that biomedical pretraining is useful for identifying causal language in PubMed-derived text. 

LLMs perform substantially lower as the best prompt-only generative model (Meta-Llama-3.3-70B) had a F$_1$ score of 0.4086, while the strongest fine-tuned generative model (Qwen-7B-FT) had a F$_1$ score of 0.3782.

\begin{table}[ht]
\centering
\caption{Reference baseline results for causal relation detection on PubMedCausal. Scores are reported for the causal class. Full prompting results are provided in Table \ref{tab:finetuned_classifiers} Appendix \ref{app:results_encoder_detection}}
\label{tab:detection_comparison}
\tiny
\begin{tabular}{lccc}
\toprule
\textbf{Model} & \textbf{Prec.} & \textbf{Rec.} & \textbf{F$_1$} \\
\midrule
\multicolumn{4}{l}{\textit{Discriminative Encoders}} \\
BERT       & 0.7249 & 0.7344 & 0.7296 \\
SciBERT    & 0.7147 & 0.7440 & 0.7291 \\
PubMedBERT & \textbf{0.7289} & \textbf{0.7496} & \textbf{0.7391} \\
BioBERT    & 0.7282 & 0.7481 & 0.7380 \\
\midrule
\multicolumn{4}{l}{\textit{Generative Decoders --- Prompt-Only (Best Configuration)}} \\
Meta-Llama-3.3-70B (ZS) & \textbf{0.3004} & 0.6386 & \textbf{0.4086} \\
DeepSeek-R1-Distill-Qwen-32B (L2M)   & 0.2264 & 0.8692 & 0.3592 \\
Mixtral-8x7B (CoT)      & 0.2266 & 0.7481 & 0.3478 \\
Llama-8B (CoT-FS)       & 0.2421 & 0.5666 & 0.3393 \\
Qwen-7B (CoT)           & 0.2122 & 0.5028 & 0.2984 \\
Mistral-7B (CoT)        & 0.1782 & 0.8697 & 0.2958 \\
Llama-3B (CoT-FS)       & 0.1825 & 0.6128 & 0.2812 \\
DeepSeek-7B (ZS)        & 0.1597 & \textbf{0.8834} & 0.2705 \\
\midrule
\multicolumn{4}{l}{\textit{Generative Decoders --- Fine-Tuned (Best Configuration)}} \\
Qwen-7B-FT (CoT)       & \textbf{0.4649} & 0.3188 & \textbf{0.3782} \\
Llama-3B-FT (CoT-FS)   & 0.2094 & \textbf{0.5692} & 0.3062 \\
Mistral-7B-FT (CoT-FS) & 0.2095 & 0.5682 & 0.3062 \\
Llama-8B-FT (FS)       & 0.1759 & 0.2899 & 0.2190 \\
DeepSeek-7B-FT (L2M)   & 0.1350 & 0.1825 & 0.1552 \\
\bottomrule
\end{tabular}
\vspace{2pt}
\\
\footnotesize ZS = Zero-Shot; FS = Few-Shot; CoT-FS = CoT Few-Shot; L2M = Least-to-Most.
\end{table}

\subsection{Experiment 2: Span-Level Causal Pair Extraction Baselines}
\label{sec:extraction_results}
In experiment 2, we extract the cause and effect pairs. In the annotation, we have observed that both cause and effect can be a singular token or an entire phrase or sentence. This is the motivation for span level extraction which covers all the scenarios. Table~\ref{tab:extraction_comparison} reports reference baselines for span-level cause-effect pair extraction on causal rows. The task requires the model to identify the full textual cause and effect spans and preserve their directionality. The task is therefore sensitive to boundary errors, missing pairs, duplicated pairs, and cause-effect reversal.

\begin{table}[ht]
\centering
\caption{For every model the best prompt strategy was reported, full prompt strategy results for all models are provided in Table~\ref{tab:bio4k_prompt_only} 
(Appendix~\ref{app:results_prompt_extraction})}
\label{tab:extraction_comparison}
\tiny
\setlength{\tabcolsep}{2.5pt}
\begin{tabular}{lccc}
\toprule
\textbf{Model} & \textbf{Token Overlap P-F$_1$} & \textbf{Exact P-F$_1$} & \textbf{Cos. P-F$_1$} \\
\midrule
\multicolumn{4}{l}{\textit{Generative Decoders --- Prompt-Only (Best Configuration)}} \\
DeepSeek-R1-Distill-Qwen-32B (FS)    & \textbf{0.5758} & \textbf{0.2383} & \textbf{0.6765} \\
Mistral-7B (L2M)        & 0.4739 & 0.1527 & 0.5958 \\
Mixtral-8x7B (FS)       & 0.3704 & 0.1469 & 0.4390 \\
Qwen-7B (FS)            & 0.3640 & 0.1400 & 0.4283 \\
Llama-8B (ReAct)        & 0.2764 & 0.1089 & 0.3191 \\
Meta-Llama-3.3-70B (ZS) & 0.2384 & 0.1013 & 0.2676 \\
Llama-3B (ReAct)        & 0.2059 & 0.0717 & 0.2554 \\
\midrule
\multicolumn{4}{l}{\textit{Generative Decoders --- Fine-Tuned (Best Configuration)}} \\
Mistral-7B-FT (L2M)     & \textbf{0.4245} & 0.1022 & \textbf{0.5812} \\
Qwen-7B-FT (ZS)         & 0.3947 & \textbf{0.1293} & 0.4817 \\
Llama-8B-FT (FS)        & 0.3175 & 0.1039 & 0.3890 \\
Llama-3B-FT (FS)        & 0.1615 & 0.0674 & 0.1919 \\
\bottomrule
\end{tabular}
\vspace{2pt}
\\
\footnotesize ZS = Zero-Shot; FS = Few-Shot; L2M = Least-to-Most; P = Pair
\end{table}

The strongest prompt-only baseline is DeepSeek-R1-Distill-Qwen-32B under few-shot prompting. Nevertheless, the observed F$_1$ scores suggest that automated causal extraction with LLMs remains far from solved, as most models perform below the 50\% threshold. These results also indicate that model size alone does not guarantee superior performance, since smaller models can outperform larger models in certain experimental settings.

Among smaller models, Mistral-7B provides the strongest base-model result under least-to-most prompting. Fine-tuned models do not consistently outperform their base counterparts: Mistral-7B drops after fine-tuning, while Qwen-7B shows marginal Token Overlap improvement but a decline on Exact Pair F$_1$, suggesting LoRA adaptation shifts rather than reliably improves extraction performance.

\subsection{Experiment 3: Transfer Learning}
\label{sec:transfer_learning}
This experiment was done to evaluate encoder models trained on PubMedCausal on selected external causal relation datasets as seen in Table~\ref{tab:selected_within_cross_F$_1$}. This is to examine whether the models trained on the corpus are capable of transferred learning. Results show stronger transfer to biomedical and general causal datasets than to financial causal text, suggesting that domain and annotation-schema differences remain important factors in cross-dataset evaluation. Full cross-dataset training parameters and dataset configurations are provided in Appendix~\ref{app:results_cross_detection} Table~\ref{tab:cross_dataset_setup}.

\definecolor{deepgray}{RGB}{70,70,70}
\definecolor{deepred}{RGB}{139,0,0}

\begin{table}[ht]
\centering
\scriptsize
\setlength{\tabcolsep}{3pt}
\tiny
\caption{Diagnostic cross-dataset causal detection results. PubMedCausal-trained models are evaluated on external datasets to estimate transfer behaviour across domains and annotation schemes.}
\label{tab:selected_within_cross_F$_1$}

\begin{tabular}{llccc}
\toprule
\textbf{Dataset} & \textbf{Model} & \textbf{Within $F_1$} & \textbf{Cross $F_1$} & \textbf{$\Delta$} \\
\midrule

\multirow{4}{*}{FinCausal}
& BERT       & 0.2113 & 0.2222 & \textcolor{deepgray}{+0.0109} \\
& SciBERT    & 0.5333 & 0.2759 & \textbf{\textcolor{deepred}{-0.2574}} \\
& PubMedBERT & 0.5098 & 0.3448 & \textcolor{deepred}{-0.1650} \\
& BioBERT    & 0.2800 & 0.2128 & \textcolor{deepred}{-0.0672} \\
\midrule

\multirow{4}{*}{CTB}
& BERT       & 0.2009 & 0.2418 & \textcolor{deepgray}{+0.0409} \\
& SciBERT    & 0.2430 & 0.1810 & \textcolor{deepred}{-0.0620} \\
& PubMedBERT & 0.2181 & 0.2015 & \textcolor{deepred}{-0.0166} \\
& BioBERT    & 0.2129 & 0.1786 & \textcolor{deepred}{-0.0343} \\
\midrule

\multirow{4}{*}{BioCause}
& BERT       & 0.3529 & 0.5556 & \textbf{\textcolor{deepgray}{+0.2027}} \\
& SciBERT    & 0.4000 & 0.5333 & \textcolor{deepgray}{+0.1333} \\
& PubMedBERT & 0.4478 & 0.5517 & \textcolor{deepgray}{+0.1039} \\
& BioBERT    & 0.5500 & 0.5000 & \textcolor{deepred}{-0.0500} \\
\midrule

\multirow{4}{*}{CRED}
& BERT       & 0.2353 & 0.1467 & \textcolor{deepred}{-0.0886} \\
& SciBERT    & 0.2545 & 0.1444 & \textcolor{deepred}{-0.1101} \\
& PubMedBERT & 0.2989 & 0.1437 & \textbf{\textcolor{deepred}{-0.1552}} \\
& BioBERT    & 0.2105 & 0.1319 & \textcolor{deepred}{-0.0786} \\
\midrule

\multirow{4}{*}{AltLex}
& BERT       & 0.6696 & 0.7525 & \textcolor{deepgray}{+0.0829} \\
& SciBERT    & 0.6332 & 0.7150 & \textcolor{deepgray}{+0.0818} \\
& PubMedBERT & 0.6723 & 0.7182 & \textcolor{deepgray}{+0.0459} \\
& BioBERT    & 0.6691 & 0.7447 & \textcolor{deepgray}{+0.0756} \\

\bottomrule
\end{tabular}

\caption*{\small
CTB = Penn Discourse Treebank-style causal text bank dataset.
CRED = Causal Relation Extraction Dataset.
Within = model fine-tuned and evaluated on the same dataset.
Cross = model trained on PubMedCausal and evaluated zero-shot on the target dataset.
$\Delta$ = Cross $F_1$ $-$ Within $F_1$.
}
\end{table}

\section{Analysis and Ablation Studies}

\subsection{Extraction Error Analysis}
\label{sec:Error Analysis}

\begin{figure}[h]
    \centering
    \includegraphics[width=0.5\textwidth]{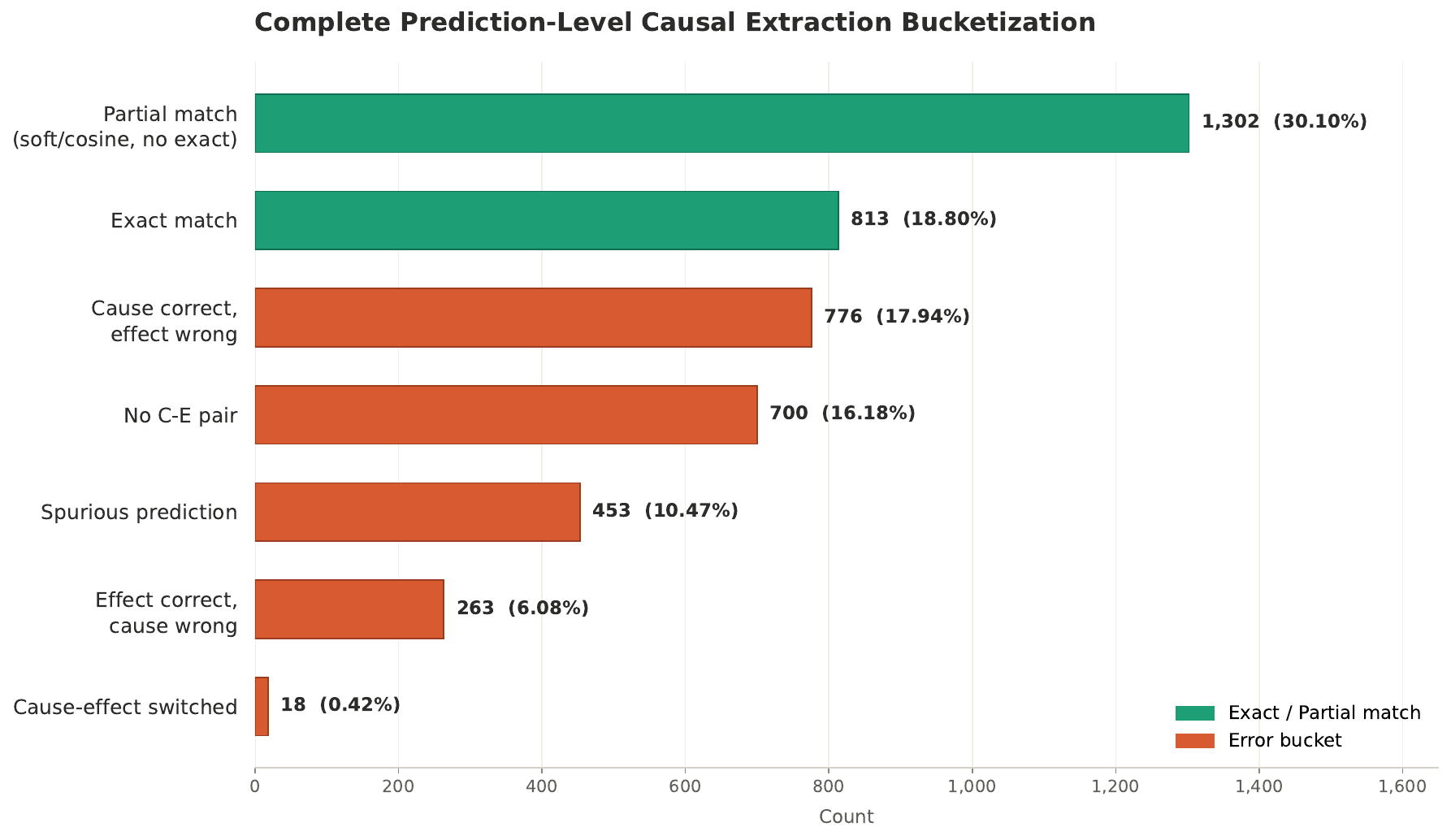}
    \caption{Error bucketization of the model/prompt strategy with the best performance}
    \label{fig:error_buckets}
\end{figure}

We conduct error analysis using the strongest extraction baseline, DeepSeek-R1-Distill-Qwen-32B with few-shot prompting. Figure~\ref{fig:error_buckets} groups predictions into error buckets indicated with a red bar. The dominant error was where the model recovers either the cause or the effect but not both accurately. This shows that PubMedCausal is challenging not only for detecting causal language, but also for recovering complete span-level causal relations. We provide full error bucket definitions and representative examples in Appendix~\ref{app:error_bucketization}.

\subsection{Detection Error Analysis}
\label{app:detection-errors}

PubMedBERT's detection errors are near-balanced in count (550 false positives, 494 false negatives; 4.2\% false-positive rate on non-causal rows) but not in kind. Rows whose relations are all implicit are recalled at 49.6\% (267/538), against 84.4\% (1{,}210/1{,}433) for rows containing at least one explicitly marked relation --- a gap of 34.8 points. Implicit-only rows are 27.3\% of causal test rows but 54.9\% of missed ones. Detection performance is therefore substantially higher when an overt causal marker is present, mirroring the extraction results in Table~\ref{tab:rq4_breakdown}. Secondary patterns are consistent: rows with an inter-sentential relation are over-represented among misses (6.3\% vs.\ 3.6\% base rate), as are rows over 85 words (8.5\% vs.\ 6.7\%), the latter potentially reflecting the 128 word-piece limit. Multi-pair rows are under-represented (32.8\% vs.\ 38.6\%).

\subsection{Performance by Causality Type and Sententiality}
\label{sec:rq4}


The experiment treated causality type and sententiality classification as an offshoot of the extraction task. Based on the best extraction model, as shown in Table~\ref{tab:rq4_breakdown}, the model performed better on direct and structurally simpler cases. Inter-sentential causality remains particularly challenging, with a baseline F$_1$ score of only 5\%, indicating substantial room for improvement.

The inter-sentential score is computed over only five exactly-matched pairs and is reported for completeness rather than as a reliable estimate; the corpus contains 202 inter-sentential relations in total, and a targeted evaluation on a rebalanced subset would be needed to measure this reliably.

In contrast, intra-sentential causality achieved an F$_1$ score of 97.43\%, showing that the model is far more effective when causal relations are expressed within a single sentence than when they must be inferred across sentence boundaries. 

In observing the causality type, we see that explicit causality achieved an F$_1$ score of 88.03\%, indicating strong performance when causal relations are clearly marked in the text. In contrast, implicit causality achieved an F$_1$ score of 39.20\%, showing that the model struggles when causality must be inferred.

\begin{table}[h]
\centering
\tiny
\begin{tabular}{lcccc}
\toprule
\textbf{Model} & \textbf{Explicit F$_1$} & \textbf{Implicit F$_1$} 
    & \textbf{Intra F$_1$} & \textbf{Inter F$_1$} \\
\midrule
DS-R1-32B (FS) & 0.8803 & 0.3920 & 0.9743 & 0.0500 \\
\bottomrule
\end{tabular}
\caption{Extraction F$_1$ broken down by causality type and sententiality. Inter-sentential and implicit causality type are consistently harder across all configurations.}
\label{tab:rq4_breakdown}
\end{table}


\subsection{Model Analysis} 
LoRA fine-tuning does not consistently improve span-level extraction: Mistral-7B performs better in its base setting, while Qwen-7B shows marginal gains on Token Overlap F$_1$ only Table~\ref{tab:small_llm_finetune_ablation}. 

The effect of causal complexity is model-dependent: Llama-3B and Llama-8B perform worse on multi-pair spans, while Mistral-7B and Qwen-7B perform better on multi-pair spans. This indicates that multi-pair extraction difficulty is not uniform across models, but depends on the interaction between model family and prompting strategy. Full ablation tables are in Appendix~\ref{app:diagnostics}.

\section{Discussion}
\label{sec:discussion}
We observe that causal detection is difficult because causal references in the real-world are not mentioned in every sentence, and as such, this huge imbalance has made it difficult for models to accurately pick up on what is causal and what is not.  In our causally biased corpus, we observe 3,945 of 30,000 instances (13.15\%) are causal, while 26,055 (86.85\%) are non-causal. Therefore, the prompting approach makes models more preemptive to explore causal statements even when none is present. The inclusion of encoder models as a first filter for causal statements could offer more practical benefits than using standalone LLMs. Even when a model correctly discovers that a sentence is causal, the ability to perform extraction requires the model to recover complete causal arguments, including long texts, implicit causality type, and multiple pairs within a single row. From our results, we see that there is still a need to develop frameworks that can handle automated extraction at a higher level of precision and recall.


PubMedCausal can therefore support future work on causal filtering, full-span extraction, implicit causality type and inter-sententiality, and pipeline-based systems that combine discriminative detection with generative extraction~\cite{zavarella-etal-2024-llm-knowledge}.

\section{Conclusion}
\label{sec:conclusion}

We introduced PubMedCausal, a 30,000-row annotated corpus for CE in biomedical text. The dataset captures span-level causal arguments, full descriptive phrases rather than atomic entities, annotated along dimensions of marker presence and sententiality. Its annotation protocol, grounded in semantics, is designed to prevent models from exploiting pre-trained domain knowledge during evaluation.

Empirical evaluation across encoder classifiers and open-source generative models shows that the dataset poses genuine difficulty. Encoder models substantially outperform generative models on binary causal detection under realistic class imbalance. Fine-tuning does not produce consistent improvements, as performance varies across prompt strategies and fine-tuned models generally do not surpass the best prompt-only baselines. The fine-tuning setup was held fixed across models to ensure comparability rather than optimized per model. We release all model configurations, prompts, and data splits to enable direct comparison, and encourage future work to build on these baselines with repeated evaluation and broader hyperparameter search.


\section*{Limitations}
Several limitations should be considered when interpreting PubMedCausal and the reported baselines.

\begin{itemize}
    \item \textbf{Corpus scope.}
    PubMedCausal is limited to English PubMed abstracts retrieved with the keyword ``causality.'' This helped us collect more causal language within the annotation budget, but it also means the corpus is not a representative sample of all biomedical writing. It may contain more abstracts on causal inference, epidemiology, and risk-factor analysis, while missing causal statements expressed through other triggers such as ``induces,'' ``reduces,'' ``contributes to,'' or ``leads to.'' As a result, performance on PubMedCausal may not directly generalize to full-text articles, clinical notes, non-English biomedical text, or other domains.

    \item \textbf{Preprocessing effects.}
    Some preprocessing choices narrowed the corpus. Rows with numbers or non-ASCII characters were removed to make annotation easier, but this may exclude causal claims involving measurements, statistical results, gene symbols, dosage information, or trial outcomes. Also, the large number of non-causal rows should be seen as a property of this query-derived corpus, not as an estimate of how common causal claims are in PubMed.

    \item \textbf{Class imbalance.}
    Explicit intra-sentential causal relations are much more common than implicit and inter-sentential relations. This makes implicit and cross-sentence causal cases harder to evaluate reliably, since fewer examples are available for those subgroups.

    \item \textbf{Annotation subjectivity.}
    Although the corpus was dual-annotated and adjudicated, causal relation extraction still involves judgment. Annotators may differ on the exact boundaries of long cause-and-effect spans. Implicit causality can also be difficult to separate from association, prediction, or background scientific discussion. PubMedCausal captures causal claims as they are written in the text; it does not verify whether those claims are biologically or clinically true.

    \item \textbf{Baseline scope.}
    The experiments are intended as reproducible reference baselines, not fully optimized systems. Hyperparameters were kept fixed across models for comparability, and results are based on controlled single-run evaluations without variance across seeds or prompt choices.

    \item \textbf{Metric limitations.}
    The automatic extraction metrics have known limits. Exact match can penalize reasonable span-boundary differences, while softer metrics may not fully capture causal direction, biomedical specificity, or output-format errors. Future work should evaluate PubMedCausal across broader biomedical subdomains, include repeated runs, expand implicit and inter-sentential cases, and test stronger model- and prompt-specific optimization.
\end{itemize}

\paragraph{Dataset Release.}
PubMedCausal is publicly available on \href{https://huggingface.co/datasets/jaypee01/PubMedCausal}{Hugging Face}, with the experimental code and evaluation scripts released on \href{https://github.com/josiahpaul07/PubMedCausal_Exp}{GitHub}. It includes the full 30{,}000-row corpus in JSON Lines format, with fields the row text, a binary causal label, and---for causal rows---a list of cause-effect pair tuples, each containing cause\_span, effect\_span, expression\_type (Explicit/Implicit), and sententiality (Intra/Inter). 
Pre-constructed train and test split indices are 
provided alongside the data. 

\bibliography{main}


\appendix

\newpage .
\newpage

\textbf{Appendix}
\section{PubMedCausal Details}
\label{app:data_details}

\begin{table}[h]
\centering
\tiny
\begin{tabular}{lrr}
\toprule
\textbf{Causal Pairs} & \textbf{Instances} & \textbf{Corpus Share (\%)} \\
\midrule
0 & 26,055 & 86.85\% \\
1 & 2,412 & 8.04\% \\
2 & 971 & 3.24\% \\
3 & 315 & 1.05\% \\
$\geq$ 4 & 247 & 0.82\% \\
\bottomrule
\end{tabular}
\caption{Distribution of cause--effect pairs per instance.}
\label{tab:pair_dist}
\end{table}

\begin{table}[h]
\centering
\tiny
\begin{tabular}{lrrr}
\toprule
\textbf{Causality Type} & \textbf{Intra-} & \textbf{Inter-} & \textbf{Total} \\
\midrule
Explicit & 4,018 & 78 & 4,096 \\
Implicit & 2,271 & 124 & 2,395 \\
\midrule
\textbf{Total} & \textbf{6,289} & \textbf{202} & \textbf{6,491} \\
\bottomrule
\end{tabular}
\caption{Distribution of extracted causal relations by sententiality and expression type.}
\label{tab:causality_types}
\end{table} 

\section{Fine-Tuning Configuration}
\label{app:finetuning_config}

\paragraph{Encoder models.} All four encoder classifiers (BERT, SciBERT, PubMedBERT, BioBERT) were fine-tuned for binary sequence classification with a learning rate of $2\times10^{-5}$, batch size 16, 3 epochs, 100 warmup steps, and a maximum sequence length of 128 word-piece tokens. Models were trained on the natural class distribution of the training split, and evaluated at the same natural distribution on the test split, with predictions taken as the argmax over class logits. The cross-dataset transfer experiments (Section~\ref{sec:transfer_learning}) use a different protocol, in which the training split is downsampled to 1:1 class balance; full settings are given in Table~\ref{tab:cross_dataset_setup}.

\paragraph{Generative models.}
Generative models were adapted with LoRA~\cite{hu-etal-2022-lora}
to reduce computational requirements while preserving task-specific
extraction behaviour. We used rank $r{=}8$ targeting only query and
value projections, reducing trainable parameters by approximately 60\%
relative to full LoRA. Restricting adaptation to $q_{\text{proj}}$ and
$v_{\text{proj}}$ is standard practice: these projections govern
attention-based retrieval and are the most impactful for adapting to
structured output formats without over-fitting. 8-bit quantisation
during training enabled larger effective batch sizes within GPU memory
constraints, while 4-bit NF4 quantisation was applied at inference for
the 32B model only, as smaller models fit in 16-bit precision.
Greedy decoding (temperature~0.0) was used throughout to ensure
reproducibility across prompt strategies. Key hyperparameters are
summarised in Table~\ref{tab:finetuning_config}.

\paragraph{Generative Model Implementation Details.}
All generative models were loaded from their official 
Hugging Face checkpoints: \texttt{meta-llama/Llama-3.2-3B-Instruct}, 
\texttt{meta-llama/Llama-3.1-8B-Instruct}, 
\texttt{mistralai/Mistral-7B-Instruct-v0.3}, 
\texttt{Qwen/Qwen2.5-7B-Instruct}, 
\texttt{deepseek-ai/DeepSeek-R1-Distill-Qwen-32B}, and 
\texttt{meta-llama/Llama-3.3-70B-Instruct}. 
Maximum input length was set to 2{,}048 tokens; maximum 
output tokens were set to 32 for detection and 256 
for extraction. Outputs were parsed by searching for 
structured tuple patterns (\texttt{Cause:}, 
\texttt{Effect:}, \texttt{Causality\_Type:}, 
\texttt{Sententiality:}) using rule-based regex; 
responses that did not contain at least one parseable 
tuple were treated as null predictions and counted as 
false negatives. Inference was run on a single NVIDIA 
A100 80GB GPU with a batch size of 1 for 32B models 
and 4 for all smaller models. Quantization used 
\texttt{bitsandbytes} 4-bit NF4 for the 32B model and 
FP16 for all others. Few-shot examples in prompts were 
drawn from the training split only; no test examples 
were used as in-context demonstrations.

\begin{table}[h]
\centering
\caption{Fine-Tuning Configuration for Generative Models.}
\label{tab:finetuning_config}
\scriptsize
\begin{tabular}{ll}
\toprule
\textbf{Parameter} & \textbf{Value} \\
\midrule
\multicolumn{2}{l}{\textit{LoRA Configuration}} \\
Rank ($r$)          & 8 \\
Alpha ($\alpha$)    & 16 \\
Target modules      & $q_{\text{proj}}$, $v_{\text{proj}}$ \\
Dropout             & 0.05 \\
\midrule
\multicolumn{2}{l}{\textit{Training Hyperparameters}} \\
Optimizer           & AdamW \\
Learning rate       & $2\times10^{-4}$ \\
LR schedule         & Cosine with 50-step warmup \\
Gradient clipping   & 1.0 \\
Per-device batch size & 2--4 (model-dependent) \\
Gradient accumulation steps & 8 \\
Effective batch size & 16--32 \\
Training epochs     & 3 \\
\midrule
\multicolumn{2}{l}{\textit{Quantisation \& Precision}} \\
Training quantisation  & 8-bit with double quantisation \\
Training precision     & FP16 \\
Inference quantisation & 4-bit NF4 (32B+ models only) \\
\bottomrule
\end{tabular}
\end{table}

\begin{table}[h]
\centering
\scriptsize
\caption{Experimental Setup for Cross-Dataset Causal Detection.}
\label{tab:cross_dataset_setup}
\setlength{\tabcolsep}{4pt}
\begin{tabular}{ll}
\toprule
\textbf{Component} & \textbf{Setting} \\
\midrule
Task                  & Binary causal detection \\
Model family          & Encoder classifiers \\
Models                & BERT, SciBERT, PubMedBERT, BioBERT \\
Split                 & 70\% train / 30\% test \\
Split strategy        & Stratified where possible \\
Training balance      & Downsample train split to 1:1 class balance \\
Test balance          & Unchanged / natural distribution \\
Validation            & Slice from balanced train split \\
Cross source          & PubMedCausal \\
Within eval.\         & Model tested on its own dataset test split \\
Cross eval.\          & PubMedCausal-trained model on other datasets \\
Seed                  & 42 \\
Decision threshold    & 0.5 \\
Max sequence length   & 256 \\
Epochs                & 3 \\
Batch size            & 16 train / 32 eval \\
Learning rate         & $2\times10^{-5}$ \\
Weight decay          & 0.01 \\
Warmup ratio          & 0.1 \\
Early stopping        & Patience $= 2$ \\
Best model metric     & F1 \\
Precision             & FP16 when CUDA is available \\
\bottomrule
\end{tabular}
\end{table}

\section{Source Data}
\subsection{Sampling Strategy, Preprocessing, and Corpus Scope}
\label{app:sampling_bias}

PubMedCausal was constructed from PubMed abstracts retrieved using the keyword ``causality'' over a defined publication period. This retrieval strategy was used as an enrichment step rather than as a representative sampling procedure. Since annotatable cause--effect relations are relatively sparse in general biomedical abstracts, a purely random PubMed sample would have required a substantially larger annotation budget to obtain enough causal instances for model development and evaluation. The keyword-based retrieval strategy therefore increased the likelihood of collecting abstracts in which causal reasoning, causal inference, or causality-related discussion might occur.

The initial PubMed search returned 24,603 abstracts. These abstracts were first segmented into sentences. Since the corpus is designed to capture both intra-sentential and inter-sentential causality, neighbouring sentences were then combined into 2--5 sentence spans. This span-based design allows the annotation to preserve local discourse context, especially where a cause is introduced in one sentence and its effect is stated or implied in a following sentence.

After span construction, we removed spans containing numbers or non-ASCII characters to improve readability and reduce annotation difficulty. This preprocessing step left 42,664 candidate spans. We initially planned to annotate 40,000 spans; however, because of annotation cost and time constraints, the final annotated corpus contains 30,000 instances. Each instance therefore consists of either a single sentence or a short paragraph-like span.

Rows containing no causal relation were retained in the corpus. This decision reflects the intended modelling setting: systems should not assume that every biomedical text span contains a cause--effect relation. Instead, they must first distinguish causal from non-causal discourse before extracting causal pairs. Retaining non-causal instances also supports causal/non-causal classification as a baseline task and makes the extraction setting more realistic than a corpus containing only positive causal examples.

\subsection{Sampling Bias and Corpus Scope}
\label{app:bias}
The PubMedCausal corpus should be interpreted as a causality-query-derived biomedical corpus rather than as a representative sample of PubMed biomedical writing. The source abstracts were retrieved from PubMed using the keyword ``causality'' over a defined time period. This retrieval strategy was chosen because the project aimed to construct a biomedical causality resource under practical annotation constraints. Since annotatable causal relations are relatively sparse in scientific abstracts, querying for causality-related papers increased the likelihood of obtaining documents in which causal reasoning, causal inference, or causal discussion might occur.

This sampling strategy introduces an important scope limitation. Abstracts retrieved using the word ``causality'' are likely to overrepresent papers concerned with causal inference, epidemiology, observational studies, methodology, risk-factor analysis, and related forms of biomedical reasoning. Conversely, the corpus may underrepresent ordinary biomedical causal statements that do not explicitly use the word ``causality'' at the abstract or metadata level. Many biomedical abstracts express causal content using verbs and constructions such as ``induces,'' ``increases,'' ``reduces,'' ``leads to,'' ``is associated with,'' or ``contributes to,'' without using the term ``causality.'' PubMedCausal should therefore not be treated as a random or fully representative PubMed corpus.

This limitation also affects how the non-causal rate should be interpreted. In the processed corpus, 26,055 of 30,000 instances contain no annotated causal relation, corresponding to 86.85\% of the data. This figure should not be read as an estimate of the prevalence of causal and non-causal sentences across PubMed as a whole. Rather, it shows that even within a corpus intentionally retrieved using a causality-oriented query, many sentences or short textual units remain descriptive, methodological, contextual, associative, or background-oriented rather than containing annotatable cause--effect relations. The high non-causal proportion is therefore informative for modelling within this query-derived setting, but it is not presented as a general statistical claim about biomedical scientific writing.

Despite this sampling bias, the corpus remains useful as a focused resource for biomedical causal relation detection and extraction. First, it provides a testbed for identifying causal relations in abstracts where causal reasoning is likely to be relevant but is not always explicitly realised at the sentence level. Second, retaining non-causal rows makes the dataset suitable for realistic causal relation detection, since models must distinguish true causal statements from surrounding scientific discourse rather than assuming that every instance contains a causal pair. Third, the corpus captures a difficult annotation setting in which causal claims are mixed with associative, evidential, and methodological language, a common feature of biomedical writing.

Future extensions of PubMedCausal could reduce this sampling limitation by adding randomly sampled PubMed abstracts, MeSH-stratified samples, disease-area-specific subsets, and retrieval queries based on broader causal trigger terms such as ``induces,'' ``causes,'' ``increases,'' ``reduces,'' ``leads to,'' and ``associated with.'' Such extensions would make it possible to compare causal language across query-derived, disease-specific, and randomly sampled biomedical corpora.

A further preprocessing limitation is that sentences containing numbers or non-ASCII characters were removed to improve annotation readability; this may exclude some quantitative biomedical causal statements involving measurements, trial outcomes, gene symbols, or statistical results.

\section{Annotation}
\subsection{Annotation Protocol}
\label{app:annotation_protocol}

The annotation task required annotators to identify directed causal relations in biomedical text. A causal relation was annotated only when the passage expressed that one event, condition, process, exposure, intervention, or state brought about, contributed to, increased, decreased, prevented, enabled, or otherwise affected another event, condition, process, or outcome. Each valid annotation therefore required a complete cause--effect pair. Passages that contained only correlation, co-occurrence, comparison, prediction, temporal succession, or general association were not annotated as causal unless the text explicitly or strongly implied a directed causal mechanism.

Annotators were instructed to read each row as a single local discourse unit. A row could contain one sentence or a short paragraph, preserving the immediate context in which causal claims appear in biomedical abstracts. If both the cause and effect occurred within the same sentence, the relation was labelled as \textit{intra-sentential}. If the cause appeared in one sentence and the effect appeared in another sentence within the same row, the relation was labelled as \textit{inter-sentential}. Relations were excluded if either argument was unavailable or could not be clearly identified from the row.

Causal arguments were expected to be explicit, meaningful spans. Pronouns, vague references, and allusions were not annotated as cause or effect spans unless their antecedents were available within the same row. For example, a sentence such as \textit{It caused hunger} was excluded if \textit{it} could not be resolved locally. However, if the previous sentence identified the antecedent, annotators resolved the pronoun and used the explicit antecedent in the cause--effect pair. Annotators were also instructed not to reject long biomedical spans when the complete cause or effect required a multi-word phrase.

Causal relations were classified as either \textit{explicit} or \textit{implicit}. Explicit causality was identified when the text contained an overt causal connective, causal verb, or causal nominal expression. Examples include \textit{cause}, \textit{because}, \textit{due to}, \textit{owing to}, \textit{lead to}, \textit{result in}, \textit{as a result}, and related forms. Implicit causality was annotated only when the causal relation was recoverable from the local context without an overt marker. Mere temporal ordering was insufficient unless the passage made the causal interpretation clear.

Annotators were explicitly warned against over-annotation. A sentence containing causal vocabulary was not automatically annotated as causal. Mentions of a possible causal relation, causal association, causal link, or the need to investigate causality were excluded when the passage did not assert a directed cause--effect relation. Reported claims were also excluded when the passage merely attributed a causal claim to another source without presenting it as an asserted finding. Similarly, speculative or hedged expressions such as \textit{may cause}, \textit{might result in}, \textit{could lead to}, or \textit{possibly increases} were excluded when the relation was presented as uncertain.

The modal \textit{can} was treated separately from speculative modals. Annotators included \textit{can} constructions when they expressed a general causal capacity or established potential effect. For example, \textit{untreated hypertension can increase stroke risk} was annotated because the statement presents hypertension as having the capacity to increase stroke risk. By contrast, \textit{hypertension may increase stroke risk} was excluded when the wording presented the relation as uncertain.

Negative causal assertions were not annotated as positive cause--effect pairs. For example, \textit{exercise did not cause mortality} was excluded because the sentence denies the causal relation. However, causally directed reduction, prevention, or protective effects were included when the passage asserted an actual effect, as in \textit{exercise reduced mortality risk} or \textit{vaccination prevented severe disease}.

\subsection{Annotation Guidelines}
\label{app:annotation_guidelines_causality_biomed}

The guidelines below were issued to human annotators
responsible for building the PubMedCausal gold standard. Annotators
were required to pass a qualification round before receiving data
batches, and their output was cross-validated to compute inter-rater
reliability. The schema (Section~2 of the box) directly defines the
four-field tuple used throughout all experiments:
\{Cause, Effect, Type, Sententiality\}.

\begin{tcolorbox}[
  title=\textbf{Annotation Guidelines: Causality in Biomedical Text},
  colback=gray!5,
  colframe=black!75,
  fonttitle=\bfseries,
  breakable
]

\subsection*{1.\ Overview and Workflow}

Annotators were asked to identify directed cause--effect relations in
biomedical text. The task focused on relations that are supported by the
provided text, rather than by external knowledge or assumptions. Each row was
treated as a local discourse unit, and annotators were instructed to extract
all valid causal relations expressed within that row.

\begin{itemize}
  \item \textbf{Batch Size:} Data was assigned in fixed-size batches.
  \item \textbf{Submission:} Completed batches were submitted for review before
        subsequent batches were assigned.
  \item \textbf{Quality Review:} Submitted annotations were checked for
        missing relations, over-annotation, span-boundary errors, incorrect
        sententiality labels, and inconsistent interpretation of the guidelines.
\end{itemize}

\subsection*{2.\ Extraction Schema}

For every identified causal relation, annotators extracted a tuple with four
fields:
\[
\{\textit{Cause},\, \textit{Effect},\, \textit{Type},\,
\textit{Sententiality}\}.
\]

\begin{description}
  \item[Cause:] The event, condition, process, exposure, intervention, or state
        that brings about or influences another outcome.
  \item[Effect:] The event, condition, process, outcome, or state that is
        brought about or influenced by the cause.
  \item[Type:] Whether the causal relation is expressed explicitly or
        implicitly.
  \item[Sententiality:] Whether the cause and effect occur within the same
        sentence or across sentence boundaries.
\end{description}

\subsection*{3.\ Inclusion and Exclusion Criteria}

\begin{itemize}
  \item Include a relation only when both the cause and the effect can be
        clearly identified from the provided row.

  \item Include causal relations expressed through clear causal predicates or
        constructions, including cases where the text states that one factor, promotes, induces, leads to,
        results in, or is the reason for another.

  \item Include \textit{can} constructions only when \textit{can} expresses a
        general causal capacity or established potential effect, such as
        \textit{``untreated hypertension can increase stroke risk.''}

  \item Exclude speculative or uncertain relations when the text presents the
        causal link as unconfirmed, hypothetical, or requiring further
        investigation, for example with forms such as \textit{may cause},
        \textit{might result in}, \textit{could lead to}, or
        \textit{possibly increases}.

  \item Exclude purely associative, correlational, predictive, comparative, or
        co-occurrence statements unless the text supports a directed causal
        interpretation.

  \item Exclude mentions of possible causal links when no direction is asserted,
        such as \textit{``the causal relationship between X and Y was
        investigated.''}

  \item Exclude reported or attributed claims when the passage merely reports
        that a causal claim was made, rather than presenting the relation as an
        asserted finding within the text.

  \item Exclude relations where either the cause or effect is vague,
        pronominal, or unavailable in the row. Pronouns may be resolved only
        when their antecedents appear within the same row.

  \item Annotate strictly from the provided text. Do not rely on external
        biomedical knowledge, external definitions, or assumptions about
        abbreviations that are not explained in the row.
\end{itemize}

\subsection*{4.\ Labelling Conventions}

\begin{itemize}
  \item \textbf{Explicit:} Use when the relation is signalled by an overt causal
        connective, causal verb, or causal nominal expression. Examples include
        \textit{because}, \textit{due to}, \textit{owing to},
        \textit{leads to}, \textit{results in}, \textit{causes},
        \textit{promotes}, \textit{reduces}, \textit{prevents}, \textit{reason}, and \textit{primary reason}.

  \item \textbf{Implicit:} Use when the causal relation is present but is not
        signalled by a dedicated causal marker. The relation must still be
        recoverable from the wording, event structure, or local discourse
        context.

  \item \textbf{Intra-sentential:} Use when both the cause and effect occur
        within the same sentence.

  \item \textbf{Inter-sentential:} Use when the cause and effect occur in
        different sentences within the same row.
\end{itemize}

\subsection*{5.\ Structural Rules}

\begin{itemize}
  \item Extract the complete meaningful span for both cause and effect. A cause
        or effect may be a single entity, a noun phrase, or a full clause,
        depending on what is needed to preserve the meaning of the relation.

  \item Decompose multiple independent effects into separate cause--effect
        pairs. For example, \textit{``A causes B and C''} should be annotated as
        two relations: \textit{A causes B} and \textit{A causes C}.

  \item If multiple factors jointly produce one effect, keep the combined cause
        as a single span when the text presents them as jointly necessary or
        jointly explanatory.

  \item Do not infer transitive causal closure. If the text states
        \textit{``A causes B, and B causes C,''} annotate \textit{A causes B}
        and \textit{B causes C}. Do not annotate \textit{A causes C} unless the
        text explicitly supports that relation.

  \item Preserve causal direction regardless of grammatical voice. For passive
        constructions such as \textit{``Y is caused by X,''} annotate
        \textbf{Cause:}~X and \textbf{Effect:}~Y.

  \item Do not annotate denied causal relations as positive cause--effect pairs.
        For example, \textit{``X did not cause Y''} should be excluded.
        However, asserted reduction or prevention relations should be included,
        as in \textit{``X reduced Y''} or \textit{``X prevented Y.''}
\end{itemize}

\subsection*{6.\ Practical Annotation Checks}

\begin{itemize}
  \item Check whether the extracted pair can be paraphrased as:
        \textit{Cause brings about Effect},
        \textit{Cause contributes to Effect}, or
        \textit{Because of Cause, Effect occurs}. If the paraphrase changes the
        meaning of the passage, revise or exclude the pair.

  \item Be careful with detection, measurement, or diagnostic statements. A test
        may reveal, detect, or identify a condition without causing that
        condition.

  \item Prioritise semantic completeness over brevity. Do not shorten spans in a
        way that removes essential biomedical meaning.

  \item Treat cue words as prompts for review, not automatic evidence of
        causality. Words such as \textit{since}, \textit{as},
        \textit{therefore}, \textit{thus}, \textit{hence}, and \textit{so} may
        signal causality in some contexts but may also express time, comparison,
        explanation, or discourse flow.

  \item When uncertain, prefer not to annotate unless the row provides enough
        linguistic evidence to identify a directed cause--effect relation.
\end{itemize}\end{tcolorbox}

\section{Prompts}
\label{app:prompts}

We evaluate six prompting strategies across both the extraction and
detection tasks. All strategies share the same underlying task
definition and causality axes; they differ only in how much
reasoning structure and in-context guidance is provided to the model.
The strategies are:

\begin{itemize}
  \item \textbf{Zero-Shot} --- task definition and rules only, no
        examples or scaffolding. This serves as the baseline for
        measuring the model's prior knowledge of causal extraction.
  \item \textbf{Few-Shot} --- adds four labelled input-output examples
        covering key edge cases (decomposition, hedging, passive voice).
  \item \textbf{Chain-of-Thought (CoT)} --- instructs the model to
        produce explicit step-by-step reasoning before each tuple,
        encouraging more deliberate modality and cardinality decisions.
  \item \textbf{Hybrid (CoT + Few-Shot)} --- combines a worked example
        with the CoT reasoning format, providing both a process
        template and a concrete demonstration.
  \item \textbf{Least-to-Most} --- decomposes the task into four
        sequential sub-steps (entity identification, pairing,
        classification, output), following the least-to-most
        prompting paradigm.
  \item \textbf{ReAct} --- frames the task as a Thought--Action--
        Observation loop, requiring the
        model to interleave reasoning and information-gathering actions
        before committing to a final answer.
\end{itemize}

\subsection{Extraction Prompts}
\label{app:extraction_prompts}

The extraction prompts ask the model to produce structured
\{Cause, Effect, Causality\_Type, Sententiality\} tuples for all
causal relations in the input. The Zero-Shot prompt below contains
the full shared task definition; subsequent prompts omit the repeated
system text and show only what is added or changed.

\label{app:pcd_zeroshot}
\begin{tcolorbox}[
  title=\textbf{Extraction: Zero-Shot Prompt},
  colback=gray!5, colframe=black!75, fonttitle=\bfseries, breakable]

\textbf{System Role:} You are an expert causal relation extraction
system. Your task is to perform Pairwise Causal Discovery (PCD) on the
provided text.

\textbf{Objective:} Extract all causal relations as tuples of
\{Cause, Effect, Causality\_Type, Sententiality\}.

\medskip
\textbf{1.\ Definition of Causality.}
Causation is the ``highest form of association.'' Distinguish strictly
between Cause (agent/trigger) and Effect (outcome). ``Can cause'' is
accepted. Hedged causal claims are not causal facts. Do not extract
relations from reported claims (e.g.\ ``Scientists believe X causes
Y''). Annotate based on the semantic structure of the text using only
common sense; do not decipher abbreviations or apply external
definitions.

\medskip
\textbf{2.\ Axis 1 — Marker Presence (Causality Type).}
\textit{Explicit} ($M{=}1$): text contains variations of ``due to,''
``results in,'' ``leads to,'' ``causes,'' or ``effect of.''
\textit{Implicit} ($M{=}0$): causality is present but none of the
above markers appear (e.g.\ ``X increased Y,'' ``X brought about Y'').

\medskip
\textbf{Axis 2 — Textual Scope (Sententiality).}
\textit{Intra-sentential} ($B{=}0$): Cause and Effect in the same
sentence. \textit{Inter-sentential} ($B{=}1$): Cause and Effect across
different sentences.

\medskip
\textbf{Axis 3 — Cardinality.}
Decompose ``A causes B and C'' into $(A{\to}B)$ and $(A{\to}C)$ when
each cause acts independently. If A and B must combine to cause C, do
not decompose.

\medskip
\textbf{No transitive closure.} If ``A causes B, which causes C,''
extract $(A{\to}B)$ and $(B{\to}C)$ only — not $(A{\to}C)$ unless
explicitly stated.

\textbf{Passive voice.} Canonicalise to active form: ``Y is caused by
X'' $\to$ Cause: X, Effect: Y.

\medskip\textbf{Input Text:}
\begin{verbatim}
[Insert Text Here]
\end{verbatim}
\textbf{Output:}
\begin{verbatim}
[Provide extracted tuples here]
\end{verbatim}
\end{tcolorbox}

\label{app:pcd_fewshot}
\begin{tcolorbox}[
  title=\textbf{Extraction: Few-Shot Prompt},
  colback=gray!5, colframe=black!75, fonttitle=\bfseries, breakable]

\textit{[Same system role, objective, and axes as Zero-Shot above.]}

\medskip
Four examples are added before the task input. They are chosen to cover
the most common annotation decisions: relation decomposition (Example~1),
hedging exclusion (Example~2), inter-sentential scope and passive voice
(Example~3), and joint causation that must not be decomposed (Example~4).

\bigskip\textbf{Examples:}

\medskip
\textit{Input:} ``Hypertension leads to heart failure and kidney damage.''

\textit{Output:}\\
Tuple 1: \{Cause: Hypertension, Effect: heart failure, Causality\_Type:
Explicit, Sententiality: Intra-sentential\}\\
Tuple 2: \{Cause: Hypertension, Effect: kidney damage, Causality\_Type:
Explicit, Sententiality: Intra-sentential\}

\medskip
\textit{Input:} ``Researchers suggest that excessive sugar intake may
cause metabolic syndrome.''

\textit{Output:} No relations extracted. (\textit{Reason:} ``suggest
that\ldots may cause'' is a reported and hedged claim.)

\medskip
\textit{Input:} ``Elevated cortisol levels attenuated immune response.
This suppression increased susceptibility to infection.''

\textit{Output:}\\
Tuple 1: \{Cause: Elevated cortisol levels, Effect: immune response
attenuation, Causality\_Type: Implicit, Sententiality:
Intra-sentential\}\\
Tuple 2: \{Cause: suppression of immune response, Effect: increased
susceptibility to infection, Causality\_Type: Implicit, Sententiality:
Inter-sentential\}

\medskip
\textit{Input:} ``The combination of alcohol and sedatives can cause
respiratory depression.''

\textit{Output:}\\
Tuple 1: \{Cause: combination of alcohol and sedatives, Effect:
respiratory depression, Causality\_Type: Explicit, Sententiality:
Intra-sentential\}

\bigskip\textbf{Task — Input Text:}
\begin{verbatim}
[Insert Text Here]
\end{verbatim}
\textbf{Output:}
\end{tcolorbox}

\label{app:pcd_cot}
\begin{tcolorbox}[
  title=\textbf{Extraction: Chain-of-Thought (CoT) Prompt},
  colback=gray!5, colframe=black!75, fonttitle=\bfseries, breakable]

\textit{[Same system role, objective, and axes as Zero-Shot above.]}

\medskip
The CoT instruction appends a six-step reasoning protocol. Unlike
Few-Shot, no labelled examples are provided; instead, the model is
asked to produce its own reasoning trace for every candidate relation
before committing to a tuple. This is intended to reduce impulsive
labelling of hedged or transitive claims.

\bigskip\textbf{Instructions.}
For every potential relationship, perform step-by-step reasoning before
outputting a tuple.

\begin{enumerate}
  \item \textbf{Identify Entities} — locate potential Cause and Effect.
  \item \textbf{Check Modality} — fact or reported claim/hedge?
        If hedge, discard.
  \item \textbf{Determine Cardinality} — does the sentence need
        decomposition?
  \item \textbf{Determine Causality Type} — check for explicit markers.
  \item \textbf{Determine Sententiality} — same sentence or across?
  \item \textbf{Formulate Output} — create the tuple.
\end{enumerate}

\textbf{Input Text:}
\begin{verbatim}
[Insert Text Here]
\end{verbatim}
\textbf{Reasoning \& Output:}
{\scriptsize
\begin{verbatim}
[Provide step-by-step reasoning then tuples]
\end{verbatim}
}
\end{tcolorbox}

\label{app:pcd_hybrid_cot_fewshot}
\begin{tcolorbox}[
  title=\textbf{Extraction: Hybrid (CoT + Few-Shot) Prompt},
  colback=gray!5, colframe=black!75, fonttitle=\bfseries, breakable]

\textit{[Same system role, objective, and axes as Zero-Shot above.]}

\medskip
This strategy combines a single fully worked example with the CoT
reasoning format. The example demonstrates how the reasoning steps map
to a real annotation decision — in particular, how passive voice and
hedging interact within the same passage.

\bigskip\textbf{Example:}

\textit{Input:} ``Tumor growth was inhibited by the new compound.
However, Dr.\ Smith claims this might cause fatigue.''

\textit{Reasoning:}\\
Sentence 1: Entities — Cause = ``new compound,'' Effect = ``Tumor
growth inhibition.'' Modality: factual. Passive voice: canonicalise.
Causality Type: ``inhibited'' not in explicit list $\Rightarrow$
Implicit. Sententiality: Intra-sentential.\\
Sentence 2: ``claims\ldots might cause'' = reported claim + hedge
$\Rightarrow$ rejected.

\textit{Final Output:}\\
Tuple 1: \{Cause: new compound, Effect: Tumor growth inhibition,
Causality\_Type: Implicit, Sententiality: Intra-sentential\}

\bigskip\textbf{Task — Input Text:}
\begin{verbatim}
[Insert Text Here]
\end{verbatim}
\textbf{Reasoning:}\\
\textbf{Final Output:}
\end{tcolorbox}

\label{app:pcd_least_to_most}
\begin{tcolorbox}[
  title=\textbf{Extraction: Least-to-Most Prompt},
  colback=gray!5, colframe=black!75, fonttitle=\bfseries, breakable]

\textit{[Same system role, objective, and axes as Zero-Shot above.]}

\medskip
Rather than reasoning within a single pass, the Least-to-Most strategy
decomposes extraction into four sequential sub-problems of increasing
specificity. This mirrors the original least-to-most prompting
framework, where solving easier sub-tasks first provides a scaffold for
harder ones such as cardinality decisions and type classification.

\bigskip
\textbf{Step 1 — Entity Identification.}
Identify all phrases acting as potential agents (Causes) or outcomes
(Effects). Ignore entities in reported claims or hedged statements.

\textbf{Step 2 — Decomposition \& Pairing.}
Pair identified entities. If a compound cause/effect exists, decide
whether they act independently (decompose) or jointly (do not
decompose).

\textbf{Step 3 — Classification.}
For each valid pair determine: (a)~Causality Type — Explicit if
``due to / results in / leads to / causes / effect of,'' otherwise
Implicit; (b)~Sententiality — Intra- vs.\ Inter-sentential.

\textbf{Step 4 — Final Extraction.}
Generate the final tuple list from Steps 1--3.

\bigskip\textbf{Input Text:}
\begin{verbatim}
[Insert Text Here]
\end{verbatim}
\end{tcolorbox}

\label{app:pcd_react}
\begin{tcolorbox}[
  title=\textbf{Extraction: ReAct (Reason + Act) Prompt},
  colback=gray!5, colframe=black!75, fonttitle=\bfseries, breakable]

\textit{[Same system role, objective, and axes as Zero-Shot above.]}

\medskip
The ReAct format frames extraction as an agent loop: each Thought
articulates a goal, each Action carries it out on the text, and each
Observation records the result. The loop is fixed to four iterations
covering scanning, filtering, canonicalisation, and classification —
corresponding to the main decision points in the annotation guidelines.

\bigskip\textbf{Instruction:} Use a Thought--Action--Observation loop.

\medskip
\textbf{Thought 1:} Scan text for causal markers or semantic causal
verbs.\\
\textbf{Action 1:} Identify candidate sentences.\\
\textbf{Observation 1:} [List candidate sentences]

\textbf{Thought 2:} Filter by modality (fact vs.\ claim/hedge).\\
\textbf{Action 2:} Remove hedged sentences; keep factual causal
statements.\\
\textbf{Observation 2:} [Filtered sentences]

\textbf{Thought 3:} Apply cardinality and passive-voice rules.\\
\textbf{Action 3:} Decompose independent ``AND'' lists; convert passive
to active.\\
\textbf{Observation 3:} [Refined pairs]

\textbf{Thought 4:} Classify Axis 1 (Explicit/Implicit) and Axis 2
(Sententiality).\\
\textbf{Action 4:} Check against explicit marker list; check sentence
boundaries.\\
\textbf{Final Answer:} Output tuples \{Cause, Effect, Causality\_Type,
Sententiality\}.

\bigskip\textbf{Input Text:}
\begin{verbatim}
[Insert Text Here]
\end{verbatim}
\end{tcolorbox}

\subsection{Detection Prompts}
\label{app:detection_prompts}

Detection is a simpler binary task: the model outputs \texttt{1} if the
passage contains at least one valid causal relation, or \texttt{0}
otherwise. The same six prompting strategies are applied. The shared
causality definition is identical to extraction, but no tuple fields
need to be produced — only the existence decision matters. This makes
detection a useful diagnostic for whether prompt strategy affects
\emph{identification} ability independently of \emph{extraction}
precision.

\label{app:causal_detection_zeroshot}
\begin{tcolorbox}[
  title=\textbf{Detection: Zero-Shot Prompt},
  colback=gray!5, colframe=black!75, fonttitle=\bfseries, breakable]

\textbf{System Role:} You are an expert causal relation detection
system. If there is even one causal instance in the text, output
\texttt{1}; if none, output \texttt{0}.

\textbf{Definition of Causality} (same principles as extraction):
strict Cause/Effect distinction; ``can cause'' is accepted; hedged and
reported claims are excluded; annotation based on semantic structure
and common sense only.

\medskip\textbf{Input Text:}
\begin{verbatim}
[Insert Text Here]
\end{verbatim}
\textbf{Output:} \texttt{"1"} or \texttt{"0"}
\end{tcolorbox}

\label{app:causal_detection_fewshot}
\begin{tcolorbox}[
  title=\textbf{Detection: Few-Shot Prompt},
  colback=gray!5, colframe=black!75, fonttitle=\bfseries, breakable]

\textit{[Same system role and definition as Detection Zero-Shot above.]}

\medskip
Three examples are provided, chosen to illustrate the two most common
failure modes in detection: false negatives from misidentified implicit
relations (Example~1 and~3) and false positives from hedged language
that superficially resembles causal claims (Example~2).

\bigskip\textbf{Examples:}

\textit{Input:} ``Hypertension leads to heart failure and kidney
damage.'' \quad \textit{Output:} \texttt{"1"}

\textit{Input:} ``Researchers suggest that excessive sugar intake may
cause metabolic syndrome.'' \quad \textit{Output:} \texttt{"0"}

\textit{Input:} ``Elevated cortisol levels attenuated immune response.
This suppression increased susceptibility to infection.''
\quad \textit{Output:} \texttt{"1"}

\bigskip\textbf{Task — Input Text:}
\begin{verbatim}
[Insert Text Here]
\end{verbatim}
\textbf{Output:}
\end{tcolorbox}

\label{app:causal_detection_cot}
\begin{tcolorbox}[
  title=\textbf{Detection: Chain-of-Thought (CoT) Prompt},
  colback=gray!5, colframe=black!75, fonttitle=\bfseries, breakable]

\textit{[Same system role and definition as Detection Zero-Shot above.]}

\medskip
For detection, the CoT protocol is condensed to three steps, since
cardinality and sententiality classification are not required. The key
reasoning bottleneck is modality: the model must correctly distinguish
between factual causal assertions and hedged or attributed claims before
issuing a verdict.

\bigskip\textbf{Instructions.}
\begin{enumerate}
  \item \textbf{Identify Entities} — locate potential Cause and Effect.
  \item \textbf{Check Modality} — fact or hedge/reported claim?
        If hedge, discard.
  \item \textbf{Output} — \texttt{"1"} if any valid causal relation
        found, \texttt{"0"} otherwise.
\end{enumerate}

\textbf{Input Text:}
\begin{verbatim}
[Insert Text Here]
\end{verbatim}
\textbf{Reasoning \& Output:}
\end{tcolorbox}

\label{app:causal_detection_hybrid_cot_fewshot}
\begin{tcolorbox}[
  title=\textbf{Detection: Hybrid (CoT + Few-Shot) Prompt},
  colback=gray!5, colframe=black!75, fonttitle=\bfseries, breakable]

\textit{[Same system role and definition as Detection Zero-Shot above.]}

\medskip
The example is chosen to be non-trivial: the passage contains both a
valid causal relation and a hedged claim in adjacent sentences.
The model must correctly apply modality filtering while recognising that
the overall verdict is \texttt{"1"} because at least one valid relation
survives after the hedge is removed.

\bigskip\textbf{Example:}

\textit{Input:} ``Tumor growth was inhibited by the new compound.
However, Dr.\ Smith claims this might cause fatigue.''

\textit{Reasoning:}\\
Sentence 1 — factual, passive; Cause = ``new compound,'' Effect =
``Tumor growth inhibition.'' Accepted.\\
Sentence 2 — ``claims\ldots might cause'' is reported + hedged.
Rejected.\\
At least one valid relation found.

\textit{Final Output:} \texttt{"1"}

\bigskip\textbf{Task — Input Text:}
\begin{verbatim}
[Insert Text Here]
\end{verbatim}
\textbf{Reasoning:}\\
\textbf{Final Output:}
\end{tcolorbox}

\label{app:causal_detection_least_to_most}
\begin{tcolorbox}[
  title=\textbf{Detection: Least-to-Most Prompt},
  colback=gray!5, colframe=black!75, fonttitle=\bfseries, breakable]

\textit{[Same system role and definition as Detection Zero-Shot above.]}

\medskip
For detection, the sub-task decomposition collapses to three steps.
Steps 1 and 2 handle identification and filtering; Step 3 aggregates
the result into a single binary verdict. This mirrors the extraction
variant but stops before classification, since type and sententiality
labels are not required.

\bigskip
\textbf{Step 1 — Entity Identification.}
Identify all phrases acting as potential agents (Causes) or outcomes
(Effects) in the text.

\textbf{Step 2 — Causal Relation Identification.}
Ignore entities in reported claims or hedged statements; accept
relations per the guidelines.

\textbf{Step 3 — Final Verdict.}
Output \texttt{"1"} if at least one valid causal relation was found;
\texttt{"0"} if none.

\bigskip\textbf{Input Text:}
\begin{verbatim}
[Insert Text Here]
\end{verbatim}
\end{tcolorbox}

\label{app:causal_detection_react}
\begin{tcolorbox}[
  title=\textbf{Detection: ReAct (Reason + Act) Prompt},
  colback=gray!5, colframe=black!75, fonttitle=\bfseries, breakable]

\textit{[Same system role and definition as Detection Zero-Shot above.]}

\medskip
The detection ReAct loop runs for three iterations rather than four,
as the canonicalisation step (passive voice, decomposition) is not
required to produce a binary output. The loop terminates after
validation against the guidelines and issues a verdict.

\bigskip\textbf{Instruction:} Use a Thought--Action--Observation loop.

\textbf{Thought 1:} Scan for causal markers or semantic causal verbs.\\
\textbf{Action 1:} Identify candidate sentences.\\
\textbf{Observation 1:} [List candidate sentences]

\textbf{Thought 2:} Filter by modality.\\
\textbf{Action 2:} Remove sentences with ``believes'' or ``suggests'';
keep factual / ``can cause'' statements.\\
\textbf{Observation 2:} [Filtered sentences]

\textbf{Thought 3:} Check remaining sentences against guidelines.\\
\textbf{Action 3:} Output \texttt{"1"} if any valid causal relation
found, \texttt{"0"} otherwise.\\
\textbf{Final Answer:} [Binary classification]

\bigskip\textbf{Input Text:}
\begin{verbatim}
[Insert Text Here]
\end{verbatim}
\end{tcolorbox}

\section{Full Experimental Results}
\label{app:full_results}

This section provides the complete numeric results for all model--
strategy combinations across both tasks. Tables are organised by task
(detection then extraction) and within each task by model type (encoder,
prompt-only generative, fine-tuned generative, cross-dataset).
Reported metrics follow the conventions defined in the main paper.

\subsection{Post-hoc Validation Selection Stability}
\label{app:val-stability}

The experiments reported in the main paper follow the original 50:50 train/test split. For the generative causal-detection baselines, all six prompting strategies were initially evaluated on the test split, and the highest-performing strategy for each model was reported in the main results, with the complete grids provided in Appendix~\ref{app:full_results}. Because this procedure uses test-set performance for strategy selection, we conducted a post-hoc check to assess whether the selected strategies were sensitive to this choice.

We sampled 1{,}000 rows from the original test split to form a validation subset and repeated prompting-strategy selection on this subset. For all eight generative models, the strategy selected on the validation subset was identical to the strategy originally selected using the full test split (Table~\ref{tab:val-selection}). This suggests that the choice of prompting strategy for causal detection was stable under the validation subsample. The validation subset is distributed with the corpus to support validation-based strategy and hyperparameter selection in future work.

\begin{table}[t]
\centering
\small
\setlength{\tabcolsep}{4pt}
\begin{tabular}{lcccc}
\toprule
\textbf{Model} &
\textbf{Test Strat.} &
\textbf{Test F$_1$} &
\textbf{Val. Strat.} &
\textbf{Val. F$_1$} \\
\midrule
Llama-3.3-70B & ZS     & 0.4086 & ZS     & 0.4316 \\
DS-R1-32B    & L2M    & 0.3592 & L2M    & 0.3672 \\
Mixtral-8x7B       & CoT    & 0.3478 & CoT    & 0.3727 \\
Llama-8B           & CoT-FS & 0.3393 & CoT-FS & 0.3164 \\
Qwen-7B            & CoT    & 0.2984 & CoT    & 0.2906 \\
Mistral-7B         & CoT    & 0.2958 & CoT    & 0.2873 \\
Llama-3B           & CoT-FS & 0.2812 & CoT-FS & 0.2939 \\
DeepSeek-7B        & ZS     & 0.2705 & ZS     & 0.2721 \\
\bottomrule
\end{tabular}
\caption{Post-hoc prompting-strategy selection stability for causal detection. Test F$_1$ denotes the originally reported score on the full 15{,}000-row test split. The validation subset contains 1{,}000 rows sampled from that test split. The highest-performing strategy on the validation subset is identical to the originally test-selected strategy for all eight models. ZS = Zero-Shot; CoT-FS = CoT Few-Shot; L2M = Least-to-Most.}
\label{tab:val-selection}
\end{table}

\subsection{Causal Detection — Encoder Models}
\label{app:results_encoder_detection}

Table~\ref{tab:finetuned_classifiers} reports the in-distribution
detection performance of the four fine-tuned encoder models on the
PubMedCausal test split. All four models achieve similar accuracy ($>0.92$)
and F1 ($\approx 0.73$), with PubMedBERT leading marginally, likely due
to its pre-training on PubMed abstracts closely matching the domain of
our corpus.

\begin{table}[h]
\small
\centering
\caption{Fine-Tuned Model Performance on Causal Detection.}
\label{tab:finetuned_classifiers}
\begin{tabular}{lcccc}
\toprule
\textbf{Model} & \textbf{Acc.} & \textbf{Prec.} & \textbf{Rec.}
               & \textbf{F1} \\
\midrule
BERT        & 0.9284 & 0.7249 & 0.7344 & 0.7296 \\
SciBERT     & 0.9273 & 0.7147 & 0.7440 & 0.7291 \\
PubMedBERT  & \textbf{0.9304} & \textbf{0.7289} & \textbf{0.7496}
            & \textbf{0.7391} \\
BioBERT     & 0.9301 & 0.7282 & 0.7481 & 0.7380 \\
\bottomrule
\end{tabular}
\end{table}

\subsection{Causal Detection — Prompt-Only Generative Models}
\label{app:results_prompt_detection}

Table~\ref{tab:prompt_only_detection} reports causal-class precision,
recall, and F1 for all eight prompt-only generative models across all
six strategies. All metrics are computed on the causal (positive) class
only, since the class imbalance makes majority-class accuracy
uninformative. Several models collapse to zero recall under structured
strategies such as CoT-FewShot and ReAct, suggesting that overly rigid
formats cause smaller models to default to the majority (non-causal)
class rather than engage with the reasoning scaffold.

\begin{table*}[h]
\centering
\small
\caption{Causal Detection --- Prompt-Only Generative Models (Causal
  Class Metrics).}
\label{tab:prompt_only_detection}
\begin{tabular}{llcccc}
\toprule
\textbf{Model} & \textbf{Strategy} & \textbf{Accuracy}
               & \textbf{Precision} & \textbf{Recall} & \textbf{F1} \\
\midrule
\multirow{6}{*}{Meta-Llama-3.3-70B}
& Zero-Shot     & 0.7568 & 0.3004 & 0.6386 & \textbf{0.4086} \\
& Few-Shot      & 0.4225 & 0.1584 & 0.7861 & 0.2637 \\
& CoT           & 0.7446 & 0.2343 & 0.4151 & 0.2995 \\
& CoT-FewShot   & 0.8682 & 0.1667 & 0.0005 & 0.0010 \\
& Least-to-Most & 0.4405 & 0.1796 & 0.9118 & 0.3001 \\
& ReAct         & 0.8680 & 0.0000 & 0.0000 & 0.0000 \\
\midrule
\multirow{6}{*}{DeepSeek-R1-Distill-Qwen-32B}
& Zero-Shot     & 0.7771 & 0.1960 & 0.2240 & 0.2091 \\
& Few-Shot      & 0.8201 & 0.1311 & 0.0654 & 0.0873 \\
& CoT           & 0.5112 & 0.2000 & 0.9057 & 0.3277 \\
& CoT-FewShot   & 0.8685 & 0.0000 & 0.0000 & 0.0000 \\
& Least-to-Most & 0.5921 & 0.2264 & 0.8692 & \textbf{0.3592} \\
& ReAct         & 0.8403 & 0.0846 & 0.0218 & 0.0347 \\
\midrule
\multirow{6}{*}{Mixtral-8x7B-Instruct}
& Zero-Shot     & 0.6124 & 0.1989 & 0.6432 & 0.3039 \\
& Few-Shot      & 0.3905 & 0.1591 & 0.8479 & 0.2679 \\
& CoT           & 0.6310 & 0.2266 & 0.7481 & 0.3478 \\
& CoT-FewShot   & 0.5535 & 0.1498 & 0.5124 & 0.2319 \\
& Least-to-Most & 0.5711 & 0.2117 & 0.8302 & 0.3374 \\
& ReAct         & 0.8684 & 0.0000 & 0.0000 & 0.0000 \\
\midrule
\multirow{6}{*}{Llama-8B}
& Zero-Shot     & 0.8089 & 0.1747 & 0.1216 & 0.1434 \\
& Few-Shot      & 0.7190 & 0.1273 & 0.1941 & 0.1538 \\
& CoT           & 0.8377 & 0.2837 & 0.1536 & 0.1993 \\
& CoT-FewShot   & 0.7097 & 0.2421 & 0.5666 & 0.3393 \\
& Least-to-Most & 0.8531 & 0.1379 & 0.0223 & 0.0384 \\
& ReAct         & 0.4702 & 0.1676 & 0.7633 & 0.2748 \\
\midrule
\multirow{6}{*}{Mistral-7B}
& Zero-Shot     & 0.4247 & 0.1612 & 0.8023 & 0.2684 \\
& Few-Shot      & 0.1459 & 0.1317 & 0.9823 & 0.2323 \\
& CoT           & 0.4553 & 0.1782 & 0.8697 & 0.2958 \\
& CoT-FewShot   & 0.8659 & 0.1833 & 0.0056 & 0.0108 \\
& Least-to-Most & 0.8715 & 0.5799 & 0.0846 & 0.1477 \\
& ReAct         & 0.2907 & 0.1512 & 0.9518 & 0.2609 \\
\midrule
\multirow{6}{*}{Llama-3B}
& Zero-Shot     & 0.8034 & 0.1772 & 0.1358 & 0.1538 \\
& Few-Shot      & 0.6576 & 0.1447 & 0.3264 & 0.2005 \\
& CoT           & 0.8675 & 0.3171 & 0.0066 & 0.0129 \\
& CoT-FewShot   & 0.5879 & 0.1825 & 0.6128 & 0.2812 \\
& Least-to-Most & 0.6993 & 0.1595 & 0.3011 & 0.2085 \\
& ReAct         & 0.7137 & 0.1537 & 0.2610 & 0.1935 \\
\midrule
\multirow{6}{*}{Qwen-7B}
& Zero-Shot     & 0.7179 & 0.2113 & 0.4187 & 0.2808 \\
& Few-Shot      & 0.5771 & 0.1412 & 0.4359 & 0.2133 \\
& CoT           & 0.6891 & 0.2122 & 0.5028 & 0.2984 \\
& CoT-FewShot   & 0.7253 & 0.1775 & 0.2995 & 0.2229 \\
& Least-to-Most & 0.8587 & 0.1524 & 0.0162 & 0.0293 \\
& ReAct         & 0.3042 & 0.1486 & 0.9067 & 0.2553 \\
\midrule
\multirow{6}{*}{DeepSeek-7B}
& Zero-Shot     & 0.3731 & 0.1597 & 0.8834 & 0.2705 \\
& Few-Shot      & 0.8685 & 0.0000 & 0.0000 & 0.0000 \\
& CoT           & 0.8685 & 0.0000 & 0.0000 & 0.0000 \\
& CoT-FewShot   & 0.8685 & 0.0000 & 0.0000 & 0.0000 \\
& Least-to-Most & 0.3151 & 0.1304 & 0.7420 & 0.2218 \\
& ReAct         & 0.1667 & 0.1351 & 0.9873 & 0.2376 \\
\bottomrule
\end{tabular}
\end{table*}

\subsection{Causal Detection — Fine-Tuned Generative Models}
\label{app:results_finetuned_detection}

Table~\ref{tab:finetuned_detection} shows detection performance after
LoRA fine-tuning. Fine-tuning generally improves precision over the
prompt-only baselines, but the benefit is uneven across strategies.
In particular, Qwen-7B-FT achieves the highest F1 of any generative
model (0.3782 under CoT), suggesting that CoT reasoning aligns well
with Qwen's fine-tuned output format. DeepSeek-7B-FT collapses to zero
recall under all strategies, indicating that fine-tuning may have
overfit the model to a narrow output pattern incompatible with its
pre-training inductive biases.

\begin{table*}[h]
\small
\centering
\caption{Causal Detection --- Fine-Tuned Generative Models (Causal
  Class Metrics).}
\label{tab:finetuned_detection}
\begin{tabular}{llcccc}
\toprule
\textbf{Model} & \textbf{Strategy} & \textbf{Accuracy}
               & \textbf{Precision} & \textbf{Recall} & \textbf{F1} \\
\midrule
\multirow{6}{*}{Qwen-7B-FT}
& Zero-Shot     & 0.8528 & 0.3468 & 0.1348 & 0.1942 \\
& Few-Shot      & 0.7466 & 0.1700 & 0.2387 & 0.1986 \\
& CoT           & 0.8621 & 0.4649 & 0.3188 & \textbf{0.3782} \\
& CoT-FewShot   & 0.8127 & 0.2977 & 0.3122 & 0.3048 \\
& Least-to-Most & 0.8315 & 0.2934 & 0.1997 & 0.2376 \\
& ReAct         & 0.7995 & 0.3101 & 0.4283 & 0.3597 \\
\midrule
\multirow{6}{*}{Llama-3B-FT}
& Zero-Shot     & 0.8097 & 0.1975 & 0.1460 & 0.1679 \\
& Few-Shot      & 0.4756 & 0.1669 & 0.7486 & 0.2730 \\
& CoT           & 0.8683 & 0.4969 & 0.0801 & 0.1379 \\
& CoT-FewShot   & 0.6607 & 0.2094 & 0.5692 & 0.3062 \\
& Least-to-Most & 0.2846 & 0.1487 & 0.9397 & 0.2568 \\
& ReAct         & 0.8554 & 0.3811 & 0.1591 & 0.2245 \\
\midrule
\multirow{6}{*}{Mistral-7B-FT}
& Zero-Shot     & 0.8097 & 0.1984 & 0.1470 & 0.1689 \\
& Few-Shot      & 0.4752 & 0.1668 & 0.7486 & 0.2729 \\
& CoT           & 0.8683 & 0.4953 & 0.0801 & 0.1379 \\
& CoT-FewShot   & 0.6613 & 0.2095 & 0.5682 & 0.3062 \\
& Least-to-Most & 0.2842 & 0.1487 & 0.9402 & 0.2568 \\
& ReAct         & 0.8553 & 0.3818 & 0.1612 & 0.2267 \\
\midrule
\multirow{6}{*}{Llama-8B-FT}
& Zero-Shot     & 0.8651 & 0.0820 & 0.0025 & 0.0049 \\
& Few-Shot      & 0.7280 & 0.1759 & 0.2899 & 0.2190 \\
& CoT           & 0.8628 & 0.2222 & 0.0172 & 0.0320 \\
& CoT-FewShot   & 0.6564 & 0.1514 & 0.3502 & 0.2114 \\
& Least-to-Most & 0.8670 & 0.0769 & 0.0010 & 0.0020 \\
& ReAct         & 0.8685 & 0.0000 & 0.0000 & 0.0000 \\
\midrule
\multirow{6}{*}{DeepSeek-7B-FT}
& Zero-Shot     & 0.8677 & 0.1333 & 0.0010 & 0.0020 \\
& Few-Shot      & 0.8529 & 0.1047 & 0.0157 & 0.0273 \\
& CoT           & 0.8683 & 0.2000 & 0.0005 & 0.0010 \\
& CoT-FewShot   & 0.8268 & 0.1257 & 0.0532 & 0.0748 \\
& Least-to-Most & 0.7387 & 0.1350 & 0.1825 & 0.1552 \\
& ReAct         & 0.8685 & 0.0000 & 0.0000 & 0.0000 \\
\bottomrule
\end{tabular}
\end{table*}

\subsection{Causal Extraction — Prompt-Only Generative Models}
\label{app:results_prompt_extraction}

Table~\ref{tab:bio4k_prompt_only} reports extraction performance using
three matching levels: Soft (token-overlap F1 on individual Cause and
Effect spans and on their pair), Exact (strict string match on the
pair), and Cosine (embedding-similarity match). Soft and Cosine metrics
are more lenient and reward semantically correct but lexically
inexact extractions; Exact Pair F1 is the strictest signal. Zeros
throughout a model's rows (e.g.\ DeepSeek-7B) indicate that the model
failed to produce parseable output in the expected format under all
six strategies.

\begin{table*}[h]
\small
\centering
\caption{Extraction Performance --- Prompt-Only Generative Models
  (Causal Pair Metrics).}
\label{tab:bio4k_prompt_only}
\setlength{\tabcolsep}{4pt}
\begin{tabular}{llcccccc}
\toprule
\textbf{Model} & \textbf{Strategy} & \textbf{Soft C-F1}
               & \textbf{Soft E-F1} & \textbf{Soft P-F1}
               & \textbf{Ex. P-F1} & \textbf{Cos. C-F1}
               & \textbf{Cos. P-F1} \\
\midrule
\multirow{6}{*}{DeepSeek-R1-Distill-Qwen-32B}
& Zero-Shot     & 0.5366 & 0.4844 & 0.5066 & 0.2222 & 0.6138 & 0.5944 \\
& Few-Shot      & \textbf{0.6166} & \textbf{0.5440} & \textbf{0.5758}
               & \textbf{0.2383} & \textbf{0.7017} & \textbf{0.6765} \\
& CoT           & 0.0000 & 0.0000 & 0.0000 & 0.0000 & 0.0000 & 0.0000 \\
& CoT-FewShot   & 0.4297 & 0.3827 & 0.4034 & 0.1811 & 0.4751 & 0.4597 \\
& Least-to-Most & 0.1114 & 0.0969 & 0.1029 & 0.0456 & 0.1176 & 0.1120 \\
& ReAct         & 0.4749 & 0.4223 & 0.4441 & 0.1775 & 0.5381 & 0.5196 \\
\midrule
\multirow{6}{*}{Mistral-7B}
& Zero-Shot     & 0.4229 & 0.4293 & 0.4197 & 0.1030 & 0.5551 & 0.5567 \\
& Few-Shot      & 0.4743 & 0.4620 & 0.4625 & 0.1428 & 0.5940 & 0.5904 \\
& CoT           & 0.4396 & 0.4303 & 0.4292 & 0.1274 & 0.5406 & 0.5380 \\
& CoT-FewShot   & 0.1809 & 0.1630 & 0.1718 & 0.0658 & 0.2090 & 0.2050 \\
& Least-to-Most & 0.4990 & 0.4607 & 0.4739 & 0.1527 & 0.6082 & \textbf{0.5958} \\
& ReAct         & 0.2207 & 0.2026 & 0.2095 & 0.0889 & 0.2558 & 0.2504 \\
\midrule
\multirow{6}{*}{Mixtral-8x7B}
& Zero-Shot     & 0.3749 & 0.3355 & 0.3527 & 0.1461 & 0.4240 & 0.4103 \\
& Few-Shot      & 0.3922 & 0.3539 & 0.3704 & 0.1469 & 0.4511 & 0.4390 \\
& CoT           & 0.1882 & 0.1742 & 0.1812 & 0.0821 & 0.2101 & 0.2063 \\
& CoT-FewShot   & 0.0000 & 0.0000 & 0.0000 & 0.0000 & 0.0000 & 0.0000 \\
& Least-to-Most & 0.1356 & 0.1257 & 0.1301 & 0.0658 & 0.1495 & 0.1462 \\
& ReAct         & 0.0000 & 0.0000 & 0.0000 & 0.0000 & 0.0000 & 0.0000 \\
\midrule
\multirow{6}{*}{Qwen-7B}
& Zero-Shot     & 0.0470 & 0.0463 & 0.0463 & 0.0171 & 0.0523 & 0.0525 \\
& Few-Shot      & 0.3897 & 0.3464 & 0.3640 & 0.1400 & 0.4429 & 0.4283 \\
& CoT           & 0.1966 & 0.1801 & 0.1871 & 0.0703 & 0.2256 & 0.2186 \\
& CoT-FewShot   & 0.1525 & 0.1362 & 0.1430 & 0.0416 & 0.1774 & 0.1714 \\
& Least-to-Most & 0.1924 & 0.1734 & 0.1815 & 0.0706 & 0.2166 & 0.2091 \\
& ReAct         & 0.2990 & 0.2798 & 0.2890 & 0.1391 & 0.3302 & 0.3232 \\
\midrule
\multirow{6}{*}{Llama-8B}
& Zero-Shot     & 0.1289 & 0.1186 & 0.1226 & 0.0471 & 0.1484 & 0.1430 \\
& Few-Shot      & 0.2848 & 0.2476 & 0.2634 & 0.0911 & 0.3214 & 0.3059 \\
& CoT           & 0.2361 & 0.2105 & 0.2200 & 0.0865 & 0.2705 & 0.2578 \\
& CoT-FewShot   & 0.0450 & 0.0405 & 0.0424 & 0.0186 & 0.0593 & 0.0575 \\
& Least-to-Most & 0.2216 & 0.1879 & 0.2030 & 0.0874 & 0.2464 & 0.2318 \\
& ReAct         & 0.2952 & 0.2636 & 0.2764 & 0.1089 & 0.3332 & 0.3191 \\
\midrule
\multirow{6}{*}{Meta-Llama-3.3-70B}
& Zero-Shot     & 0.2581 & 0.2199 & 0.2384 & 0.1013 & 0.2810 & 0.2676 \\
& Few-Shot      & 0.0250 & 0.0217 & 0.0232 & 0.0086 & 0.0280 & 0.0270 \\
& CoT           & 0.0000 & 0.0000 & 0.0000 & 0.0000 & 0.0000 & 0.0000 \\
& CoT-FewShot   & 0.0019 & 0.0014 & 0.0016 & 0.0006 & 0.0019 & 0.0017 \\
& Least-to-Most & 0.0000 & 0.0000 & 0.0000 & 0.0000 & 0.0000 & 0.0000 \\
& ReAct         & 0.0006 & 0.0006 & 0.0006 & 0.0006 & 0.0006 & 0.0006 \\
\midrule
\multirow{6}{*}{Llama-3B}
& Zero-Shot     & 0.1494 & 0.1224 & 0.1341 & 0.0526 & 0.1725 & 0.1606 \\
& Few-Shot      & 0.0469 & 0.0429 & 0.0436 & 0.0072 & 0.0579 & 0.0550 \\
& CoT           & 0.1228 & 0.1129 & 0.1166 & 0.0481 & 0.1388 & 0.1343 \\
& CoT-FewShot   & 0.1713 & 0.1529 & 0.1606 & 0.0601 & 0.1965 & 0.1885 \\
& Least-to-Most & 0.0028 & 0.0022 & 0.0025 & 0.0012 & 0.0030 & 0.0027 \\
& ReAct         & 0.2162 & 0.2007 & 0.2059 & 0.0717 & 0.2637 & 0.2554 \\
\midrule
\multirow{6}{*}{DeepSeek-7B}
& Zero-Shot     & 0.0000 & 0.0000 & 0.0000 & 0.0000 & 0.0000 & 0.0000 \\
& Few-Shot      & 0.0000 & 0.0000 & 0.0000 & 0.0000 & 0.0000 & 0.0000 \\
& CoT           & 0.0000 & 0.0000 & 0.0000 & 0.0000 & 0.0000 & 0.0000 \\
& CoT-FewShot   & 0.0000 & 0.0000 & 0.0000 & 0.0000 & 0.0000 & 0.0000 \\
& Least-to-Most & 0.0000 & 0.0000 & 0.0000 & 0.0000 & 0.0000 & 0.0000 \\
& ReAct         & 0.0000 & 0.0000 & 0.0000 & 0.0000 & 0.0000 & 0.0000 \\
\bottomrule
\end{tabular}
\end{table*}

\subsection{Causal Extraction — Fine-Tuned Generative Models}
\label{app:results_finetuned_extraction}

Table~\ref{tab:bio4k_finetuned} reports extraction F1 after LoRA
fine-tuning. Mistral-7B-FT with Least-to-Most achieves the strongest
overall extraction performance (Soft P-F1 = 0.4245; Cos. P-F1 = 0.5812),
suggesting that the structured sub-task decomposition is particularly
compatible with Mistral's instruction-following tendencies after
fine-tuning. As with detection, DeepSeek-7B-FT produces zero output
under all strategies, confirming a persistent format-alignment failure
that fine-tuning did not resolve.

\begin{table*}[h]
\small
\centering
\caption{Extraction Performance --- Fine-Tuned Generative Models
  (Causal Pair Metrics).}
\label{tab:bio4k_finetuned}
\begin{tabular}{llcccccc}
\toprule
\textbf{Model} & \textbf{Strategy} & \textbf{Soft C-F1}
               & \textbf{Soft E-F1} & \textbf{Soft P-F1}
               & \textbf{Ex. P-F1} & \textbf{Cos. C-F1}
               & \textbf{Cos. P-F1} \\
\midrule
\multirow{6}{*}{Mistral-7B-FT}
& Zero-Shot     & 0.3224 & 0.3313 & 0.3240 & 0.0756 & 0.4873 & 0.4883 \\
& Few-Shot      & 0.0171 & 0.0229 & 0.0199 & 0.0054 & 0.0328 & 0.0345 \\
& CoT           & 0.3380 & \textbf{0.3601} & 0.3469 & 0.0965 & 0.5001 & 0.5109 \\
& CoT-FewShot   & 0.0047 & 0.0035 & 0.0040 & 0.0012 & 0.0060 & 0.0056 \\
& Least-to-Most & \textbf{0.4322} & 0.4235 & \textbf{0.4245} & 0.1022
               & \textbf{0.5827} & \textbf{0.5812} \\
& ReAct         & 0.0991 & 0.1145 & 0.1059 & 0.0256 & 0.2558 & 0.2594 \\
\midrule
\multirow{6}{*}{Qwen-7B-FT}
& Zero-Shot     & 0.4137 & 0.3856 & 0.3947 & \textbf{0.1293} & 0.4937 & 0.4817 \\
& Few-Shot      & 0.3689 & 0.3266 & 0.3439 & 0.1287 & 0.4244 & 0.4091 \\
& CoT           & 0.1476 & 0.1312 & 0.1394 & 0.0766 & 0.1612 & 0.1557 \\
& CoT-FewShot   & 0.2726 & 0.2592 & 0.2645 & 0.1220 & 0.3014 & 0.2962 \\
& Least-to-Most & 0.1086 & 0.0996 & 0.1041 & 0.0678 & 0.1138 & 0.1112 \\
& ReAct         & 0.0000 & 0.0000 & 0.0000 & 0.0000 & 0.0000 & 0.0000 \\
\midrule
\multirow{6}{*}{Llama-8B-FT}
& Zero-Shot     & 0.0216 & 0.0216 & 0.0216 & 0.0080 & 0.0269 & 0.0269 \\
& Few-Shot      & 0.3348 & 0.3086 & 0.3175 & 0.1039 & 0.4015 & 0.3890 \\
& CoT           & 0.2665 & 0.2530 & 0.2578 & 0.0697 & 0.3263 & 0.3192 \\
& CoT-FewShot   & 0.1695 & 0.1661 & 0.1666 & 0.0596 & 0.2179 & 0.2159 \\
& Least-to-Most & 0.0124 & 0.0131 & 0.0126 & 0.0043 & 0.0143 & 0.0142 \\
& ReAct         & 0.0006 & 0.0006 & 0.0006 & 0.0006 & 0.0006 & 0.0006 \\
\midrule
\multirow{6}{*}{Llama-3B-FT}
& Zero-Shot     & 0.1597 & 0.1407 & 0.1477 & 0.0611 & 0.1802 & 0.1731 \\
& Few-Shot      & 0.1748 & 0.1496 & 0.1615 & 0.0674 & 0.2002 & 0.1919 \\
& CoT           & 0.0568 & 0.0534 & 0.0547 & 0.0204 & 0.0701 & 0.0690 \\
& CoT-FewShot   & 0.0482 & 0.0436 & 0.0458 & 0.0234 & 0.0607 & 0.0604 \\
& Least-to-Most & 0.0017 & 0.0014 & 0.0016 & 0.0006 & 0.0019 & 0.0018 \\
& ReAct         & 0.0091 & 0.0083 & 0.0087 & 0.0019 & 0.0110 & 0.0107 \\
\midrule
\multirow{6}{*}{DeepSeek-7B-FT}
& Zero-Shot     & 0.0000 & 0.0000 & 0.0000 & 0.0000 & 0.0000 & 0.0000 \\
& Few-Shot      & 0.0000 & 0.0000 & 0.0000 & 0.0000 & 0.0000 & 0.0000 \\
& CoT           & 0.0000 & 0.0000 & 0.0000 & 0.0000 & 0.0000 & 0.0000 \\
& CoT-FewShot   & 0.0000 & 0.0000 & 0.0000 & 0.0000 & 0.0000 & 0.0000 \\
& Least-to-Most & 0.0000 & 0.0000 & 0.0000 & 0.0000 & 0.0000 & 0.0000 \\
& ReAct         & 0.0000 & 0.0000 & 0.0000 & 0.0000 & 0.0000 & 0.0000 \\
\bottomrule
\end{tabular}
\end{table*}

\section{Cross-Dataset Transfer Detection Setup and Results}
\label{app:results_cross_detection}

This appendix provides the full setup for the cross-dataset causal detection experiment referenced in Section~\ref{sec:transfer_learning}. 
The goal of this experiment is to assess whether encoder models trained on PubMedCausal transfer to external causal relation datasets with different domains and annotation schemes. 
For each dataset, models were first fine-tuned and evaluated on the same dataset to obtain a within-dataset baseline. 
The cross-dataset setting then evaluated models trained on PubMedCausal directly on the held-out test splits of external datasets without additional fine-tuning on the target dataset.

The transfer results show that PubMedCausal-trained encoders transfer more strongly to BioCause and AltLex than to FinCausal and CTB. 
The largest positive transfer is observed for BERT on BioCause, where cross-dataset $F_1$ improves by $+0.2027$ over the within-dataset baseline. 
The largest negative transfer is observed for SciBERT on FinCausal, with a drop of $-0.2574$. 
This pattern suggests that transfer is sensitive to domain and annotation-schema alignment: biomedical and general causal datasets benefit more from PubMedCausal training, while financial causal text remains less compatible with the learned decision boundary.

\subsection{Cross-Dataset Causal Extraction}
\label{app:results_cross_extraction}

Tables~\ref{tab:causenet_base_extraction}--\ref{tab:biocause_base_extraction}
report extraction performance of base (non-fine-tuned) models on three
out-of-distribution datasets. Each table shows Exact, Token-overlap,
and Cosine F1 for Cause, Effect, and Pair. The consistent gap between
Cosine and Exact F1 across all models and datasets indicates that
models frequently recover the correct semantic content but use different
surface phrasing than the reference annotation — a known challenge in
span-level extraction benchmarks. Tables~\ref{tab:causenet_results}
through~\ref{tab:biocause_results} show the same breakdown after
fine-tuning on PubMedCausal; Llama-8B consistently achieves the
strongest generalisation across all three external datasets.

\subsection{Error Bucket Definitions}
\label{app:error_bucketization}

Figure~\ref{fig:error_buckets} reports the prediction-level error bucketization for the best-performing causal extraction configuration, DeepSeek-R1-Distill-Qwen-32B with few-shot prompting. 
The analysis is conducted at the cause--effect pair level rather than the sentence level. 
Each predicted pair is aligned to the closest gold pair within the same input span and assigned to one mutually exclusive bucket using a priority-based scheme. 
Exact and partial matches are separated from error categories, while the remaining buckets describe the dominant structural failure modes observed in the model outputs.

\begin{table*}[t]
\centering
\small
\setlength{\tabcolsep}{3pt}
\caption{Definitions and examples of prediction-level error bucketization categories. Examples are drawn from DeepSeek-R1-Distill-Qwen-32B Few-shot predictions and the corresponding gold annotations.}
\label{tab:error_bucket_definitions_examples}
\begin{tabular}{p{5.0cm}p{5.2cm}p{5.2cm}}
\toprule
\textbf{Definition} & \textbf{Prediction} & \textbf{Gold} \\
\midrule

\textbf{Exact match:} The predicted cause and effect spans exactly match the gold cause and effect after normalization.
& Prevotellaceae $\rightarrow$ sepsis
& Prevotellaceae $\rightarrow$ sepsis \\

\midrule

\textbf{Partial match:} The prediction is not an exact span match, but it preserves meaningful lexical or semantic overlap with the gold pair under soft-token overlap or cosine similarity.
& higher insulin levels $\rightarrow$ endometrial cancer risk
& high insulin level $\rightarrow$ an increased risk of endometrial cancer \\

\midrule

\textbf{Cause correct, effect wrong:} The predicted cause matches the gold cause, but the predicted effect is incorrect, incomplete, or aligned to the wrong outcome.
& heterogeneity of both built environment and QoL measures $\rightarrow$ consistent definitions of both concepts
& heterogeneity of both built environment and QoL measures $\rightarrow$ consistent definitions of both concepts will help advance this area of research \\

\midrule

\textbf{Effect correct, cause wrong:} The predicted effect matches the gold effect, but the predicted cause is incorrect or unsupported by the gold annotation.
& AIP $\rightarrow$ AR-HCC
& PBGD $\rightarrow$ AR-HCC \\

\midrule

\textbf{Switched relation:} The model reverses the causal direction, such that the predicted cause corresponds to the gold effect and the predicted effect corresponds to the gold cause.
& chronic widespread pain $\rightarrow$ GERD
& GERD $\rightarrow$ chronic widespread pain \\

\midrule

\textbf{Spurious prediction:} The model predicts a cause--effect pair that does not reasonably correspond to any gold causal pair in the input.
& ADR $\rightarrow$ diarrhea
& Antimicrobials $\rightarrow$ ADRs \\

\midrule

\textbf{No C-E pair:} A gold cause--effect pair exists, but the model produces no causal pair for that sentence or span.
& No prediction
& the cross-sectional design $\rightarrow$ causal inference is limited \\

\bottomrule
\end{tabular}
\end{table*}

The distinction between exact and partial matches is important because biomedical causal spans are often long and non-atomic. 
A prediction may preserve the core causal meaning while differing from the gold annotation in span boundaries. 
For example, the prediction ``higher insulin levels'' $\rightarrow$ ``endometrial cancer risk'' is not an exact match for the gold pair ``high insulin level'' $\rightarrow$ ``an increased risk of endometrial cancer'', but it retains the same causal content and is therefore treated as a partial match. 
The remaining buckets capture genuine extraction failures. 
Argument-level errors occur when one side of the relation is correct but the other is wrong, as in cause-correct/effect-wrong or effect-correct/cause-wrong cases. 
Switched relations capture directionality errors, where the model reverses the causal arrow. 
Spurious predictions indicate unsupported causal pairs generated by the model, while No C-E pair errors indicate annotated gold relations for which the model produced no extraction. 
This bucketization therefore separates boundary-level variation from more substantive causal extraction failures Table\ref{tab:error_bucket_definitions_examples}.

\begin{table}[h]
\centering
\caption{CauseNet Extraction --- Base Models.}
\small
\label{tab:causenet_base_extraction}
\begin{tabular}{llccc}
\toprule
\textbf{Model} & \textbf{Metric} & \textbf{Cause F1}
               & \textbf{Effect F1} & \textbf{Pair F1} \\
\midrule
\multirow{3}{*}{Llama-3.2-3B}
& Exact  & 0.1689 & 0.1948 & 0.0455 \\
& Token  & 0.1514 & 0.1584 & 0.1546 \\
& Cosine & 0.3311 & 0.3224 & 0.3263 \\
\midrule
\multirow{3}{*}{Llama-3.1-8B}
& Exact  & 0.1457 & 0.2333 & 0.0518 \\
& Token  & 0.1959 & 0.2554 & 0.2265 \\
& Cosine & 0.5068 & 0.5063 & 0.5065 \\
\midrule
\multirow{3}{*}{Mistral-7B}
& Exact  & 0.2082 & 0.1870 & 0.0593 \\
& Token  & 0.2726 & 0.2605 & 0.2663 \\
& Cosine & 0.6222 & 0.6003 & 0.6091 \\
\midrule
\multirow{3}{*}{Qwen2.5-7B}
& Exact  & 0.0153 & 0.0830 & 0.0060 \\
& Token  & 0.0219 & 0.0436 & 0.0346 \\
& Cosine & 0.0686 & 0.0722 & 0.0685 \\
\bottomrule
\end{tabular}
\end{table}

\begin{table}[h]
\centering
\caption{BioCause Extraction --- Base Models.}
\small
\label{tab:biocause_base_extraction}
\begin{tabular}{llccc}
\toprule
\textbf{Model} & \textbf{Metric} & \textbf{Cause F1}
               & \textbf{Effect F1} & \textbf{Pair F1} \\
\midrule
\multirow{3}{*}{Llama-3.2-3B}
& Exact  & 0.0417 & 0.0774 & 0.0124 \\
& Token  & 0.2821 & 0.1535 & 0.2219 \\
& Cosine & 0.3540 & 0.2408 & 0.2998 \\
\midrule
\multirow{3}{*}{Llama-3.1-8B}
& Exact  & 0.0796 & 0.0939 & 0.0254 \\
& Token  & 0.2847 & 0.1884 & 0.2386 \\
& Cosine & 0.3438 & 0.2699 & 0.3083 \\
\midrule
\multirow{3}{*}{Mistral-7B}
& Exact  & 0.0485 & 0.0469 & 0.0215 \\
& Token  & 0.1602 & 0.1553 & 0.1569 \\
& Cosine & 0.2333 & 0.2335 & 0.2326 \\
\midrule
\multirow{3}{*}{Qwen2.5-7B}
& Exact  & 0.0000 & 0.0351 & 0.0000 \\
& Token  & 0.0250 & 0.0351 & 0.0305 \\
& Cosine & 0.0314 & 0.0369 & 0.0342 \\
\bottomrule
\end{tabular}
\end{table}

\begin{table}[h]
\centering
\caption{Coling22-Mining Extraction --- Base Models.}
\small
\label{tab:coling22_base_extraction}
\begin{tabular}{llccc}
\toprule
\textbf{Model} & \textbf{Metric} & \textbf{Cause F1}
               & \textbf{Effect F1} & \textbf{Pair F1} \\
\midrule
\multirow{3}{*}{Llama-3.2-3B}
& Exact  & 0.1858 & 0.2014 & 0.0961 \\
& Token  & 0.2963 & 0.2802 & 0.2890 \\
& Cosine & 0.3538 & 0.3406 & 0.3462 \\
\midrule
\multirow{3}{*}{Llama-3.1-8B}
& Exact  & 0.3133 & 0.2901 & 0.1491 \\
& Token  & 0.4591 & 0.4490 & 0.4518 \\
& Cosine & 0.5364 & 0.5263 & 0.5282 \\
\midrule
\multirow{3}{*}{Mistral-7B}
& Exact  & 0.3398 & 0.2945 & 0.1739 \\
& Token  & 0.5718 & 0.5216 & 0.5445 \\
& Cosine & 0.6803 & 0.6424 & 0.6581 \\
\midrule
\multirow{3}{*}{Qwen2.5-7B}
& Exact  & 0.0535 & 0.0563 & 0.0249 \\
& Token  & 0.0889 & 0.0845 & 0.0877 \\
& Cosine & 0.1035 & 0.0896 & 0.0968 \\
\bottomrule
\end{tabular}
\end{table}

\begin{table*}[h]
\centering
\small
\caption{CauseNet Dataset --- Extraction Performance by Fine-Tuned
  Model.}
\label{tab:causenet_results}
\begin{tabular}{llccccccccc}
\toprule
\textbf{Model} & \textbf{Metric} &
\multicolumn{3}{c}{\textbf{Cause}} &
\multicolumn{3}{c}{\textbf{Effect}} &
\multicolumn{3}{c}{\textbf{Pair}} \\
\cmidrule(lr){3-5}\cmidrule(lr){6-8}\cmidrule(lr){9-11}
& & P & R & F1 & P & R & F1 & P & R & F1 \\
\midrule
\multirow{3}{*}{Llama-3B}
& Exact  & 0.1838 & 0.1447 & 0.1619 & 0.2105 & 0.2116 & 0.2111 & 0.0582 & 0.0597 & 0.0589 \\
& Token  & 0.2183 & 0.1421 & 0.1721 & 0.2566 & 0.1662 & 0.2017 & 0.2374 & 0.1523 & 0.1855 \\
& Cosine & 0.5887 & 0.2722 & 0.3723 & 0.5340 & 0.2702 & 0.3588 & 0.5614 & 0.2694 & 0.3641 \\
\midrule
\multirow{3}{*}{Llama-8B}
& Exact  & 0.1818 & 0.2298 & 0.2030 & 0.2007 & 0.3016 & 0.2410 & 0.0659 & 0.0896 & 0.0759 \\
& Token  & 0.2846 & 0.2976 & 0.2909 & 0.2789 & 0.3284 & 0.3016 & 0.2818 & 0.3130 & 0.2966 \\
& Cosine & 0.6268 & 0.5949 & 0.6104 & 0.5971 & 0.6018 & 0.5995 & 0.6120 & 0.5980 & 0.6049 \\
\midrule
\multirow{3}{*}{Mistral-7B}
& Exact  & 0.1872 & 0.1489 & 0.1659 & 0.1139 & 0.1217 & 0.1176 & 0.0323 & 0.0261 & 0.0289 \\
& Token  & 0.2600 & 0.1841 & 0.2156 & 0.1562 & 0.1186 & 0.1348 & 0.2081 & 0.1514 & 0.1752 \\
& Cosine & 0.5714 & 0.3661 & 0.4463 & 0.5024 & 0.3272 & 0.3963 & 0.5369 & 0.3439 & 0.4192 \\
\midrule
\multirow{3}{*}{Qwen-7B}
& Exact  & 0.2000 & 0.0255 & 0.0453 & 0.2069 & 0.0635 & 0.0972 & 0.0147 & 0.0037 & 0.0060 \\
& Token  & 0.2856 & 0.0183 & 0.0343 & 0.1265 & 0.0315 & 0.0504 & 0.2061 & 0.0249 & 0.0444 \\
& Cosine & 0.6617 & 0.0452 & 0.0847 & 0.4117 & 0.0466 & 0.0838 & 0.5367 & 0.0459 & 0.0846 \\
\bottomrule
\end{tabular}
\end{table*}

\begin{table*}[h]
\centering
\small
\caption{Coling22-Mining Dataset --- Extraction Performance by
  Fine-Tuned Model.}
\label{tab:coling22_results}
\begin{tabular}{llccccccccc}
\toprule
\textbf{Model} & \textbf{Metric} &
\multicolumn{3}{c}{\textbf{Cause}} &
\multicolumn{3}{c}{\textbf{Effect}} &
\multicolumn{3}{c}{\textbf{Pair}} \\
\cmidrule(lr){3-5}\cmidrule(lr){6-8}\cmidrule(lr){9-11}
& & P & R & F1 & P & R & F1 & P & R & F1 \\
\midrule
\multirow{3}{*}{Llama-3B}
& Exact  & 0.2756 & 0.1581 & 0.2009 & 0.2564 & 0.1471 & 0.1869 & 0.1183 & 0.0806 & 0.0959 \\
& Token  & 0.4569 & 0.2201 & 0.2971 & 0.4476 & 0.2244 & 0.2989 & 0.4522 & 0.2198 & 0.2958 \\
& Cosine & 0.5872 & 0.2510 & 0.3517 & 0.5794 & 0.2552 & 0.3543 & 0.5833 & 0.2505 & 0.3505 \\
\midrule
\multirow{3}{*}{Llama-8B}
& Exact  & 0.3250 & 0.3824 & 0.3514 & 0.3047 & 0.3787 & 0.3377 & 0.1506 & 0.2125 & 0.1763 \\
& Token  & 0.5360 & 0.5743 & 0.5545 & 0.5160 & 0.5876 & 0.5495 & 0.5260 & 0.5783 & 0.5509 \\
& Cosine & 0.6477 & 0.6539 & 0.6508 & 0.6426 & 0.6711 & 0.6565 & 0.6452 & 0.6600 & 0.6525 \\
\midrule
\multirow{3}{*}{Mistral-7B}
& Exact  & 0.2452 & 0.1397 & 0.1780 & 0.1950 & 0.1140 & 0.1439 & 0.0732 & 0.0440 & 0.0549 \\
& Token  & 0.4167 & 0.2295 & 0.2960 & 0.3970 & 0.2266 & 0.2885 & 0.4068 & 0.2263 & 0.2908 \\
& Cosine & 0.5779 & 0.3014 & 0.3961 & 0.5771 & 0.3018 & 0.3963 & 0.5775 & 0.2986 & 0.3937 \\
\midrule
\multirow{3}{*}{Qwen-7B}
& Exact  & 0.3333 & 0.0404 & 0.0721 & 0.1739 & 0.0294 & 0.0503 & 0.0851 & 0.0147 & 0.0250 \\
& Token  & 0.5305 & 0.0553 & 0.1001 & 0.3055 & 0.0491 & 0.0847 & 0.4180 & 0.0520 & 0.0925 \\
& Cosine & 0.6480 & 0.0633 & 0.1154 & 0.4541 & 0.0578 & 0.1025 & 0.5510 & 0.0599 & 0.1081 \\
\bottomrule
\end{tabular}
\end{table*}

\begin{table*}[h]
\centering
\small
\caption{BioCause Dataset --- Extraction Performance by Fine-Tuned
  Model.}
\label{tab:biocause_results}
\begin{tabular}{llccccccccc}
\toprule
\textbf{Model} & \textbf{Metric} &
\multicolumn{3}{c}{\textbf{Cause}} &
\multicolumn{3}{c}{\textbf{Effect}} &
\multicolumn{3}{c}{\textbf{Pair}} \\
\cmidrule(lr){3-5}\cmidrule(lr){6-8}\cmidrule(lr){9-11}
& & P & R & F1 & P & R & F1 & P & R & F1 \\
\midrule
\multirow{3}{*}{Llama-3B}
& Exact  & 0.0087 & 0.0227 & 0.0126 & 0.0413 & 0.0980 & 0.0581 & 0.0000 & 0.0000 & 0.0000 \\
& Token  & 0.2137 & 0.2097 & 0.2117 & 0.1003 & 0.2538 & 0.1438 & 0.1570 & 0.2268 & 0.1856 \\
& Cosine & 0.2635 & 0.2742 & 0.2687 & 0.1547 & 0.3102 & 0.2065 & 0.2091 & 0.2907 & 0.2432 \\
\midrule
\multirow{3}{*}{Llama-8B}
& Exact  & 0.0130 & 0.0682 & 0.0219 & 0.0290 & 0.1373 & 0.0479 & 0.0079 & 0.0392 & 0.0132 \\
& Token  & 0.1794 & 0.4029 & 0.2483 & 0.0826 & 0.3656 & 0.1348 & 0.1310 & 0.3778 & 0.1945 \\
& Cosine & 0.2316 & 0.5759 & 0.3304 & 0.1427 & 0.5345 & 0.2252 & 0.1871 & 0.5465 & 0.2788 \\
\midrule
\multirow{3}{*}{Mistral-7B}
& Exact  & 0.0065 & 0.0227 & 0.0102 & 0.0000 & 0.0000 & 0.0000 & 0.0000 & 0.0000 & 0.0000 \\
& Token  & 0.1467 & 0.2482 & 0.1844 & 0.0826 & 0.2387 & 0.1228 & 0.1146 & 0.2250 & 0.1519 \\
& Cosine & 0.2332 & 0.3581 & 0.2825 & 0.1654 & 0.3896 & 0.2322 & 0.1993 & 0.3558 & 0.2555 \\
\midrule
\multirow{3}{*}{Qwen-7B}
& Exact  & 0.0909 & 0.0227 & 0.0364 & 0.3529 & 0.1176 & 0.1765 & 0.2353 & 0.0784 & 0.1176 \\
& Token  & 0.8464 & 0.1709 & 0.2843 & 0.2538 & 0.1660 & 0.2007 & 0.5501 & 0.1684 & 0.2579 \\
& Cosine & 0.8757 & 0.1869 & 0.3080 & 0.3957 & 0.1847 & 0.2519 & 0.6357 & 0.1858 & 0.2875 \\
\bottomrule
\end{tabular}
\end{table*}

\section{Additional Diagnostic Experiments}
\label{app:diagnostics}
To better understand model behaviour beyond aggregate extraction scores, we report two compact ablation analyses. First, we compare prompt-only and fine-tuned generative models to test whether LoRA adaptation improves span-level extraction. Also, we separate single-pair and multi-pair causal spans to examine the effect of causal cardinality. 

\begin{table}[t]
\centering
\tiny
\setlength{\tabcolsep}{3pt}
\caption{Fine-tuning ablation for the two strongest small LLMs in causal extraction.}
\label{tab:small_llm_finetune_ablation}
\begin{tabular}{llccc}
\toprule
\textbf{Model} & \textbf{Setting} & \textbf{Token Overlap} & \textbf{Exact} & \textbf{Cos.} \\
\midrule
Mistral-7B & Base L2M & 0.4739 & 0.1527 & 0.5958 \\
Mistral-7B & FT L2M & 0.4245 & 0.1022 & 0.5812 \\
\midrule
Qwen-7B & Base FS & 0.3640 & 0.1400 & 0.4283 \\
Qwen-7B & FT ZS & 0.3947 & 0.1293 & 0.4817 \\
\bottomrule
\end{tabular}
\end{table}

\begin{table}[t]
\centering
\small
\setlength{\tabcolsep}{3pt}
\caption{Small-LLM cardinality ablation.}
\label{tab:small_llm_cardinality_ablation}
\begin{tabular}{lccc}
\toprule
\textbf{Model} & \textbf{Single} & \textbf{Multi} & \textbf{Comb.} \\
\midrule
Llama-3B & 0.0492 & 0.0131 & 0.0717 \\
Llama-8B & 0.0859 & 0.0376 & 0.1089 \\
Mistral-7B & 0.1044 & 0.1777 & 0.1527 \\
Qwen-7B & 0.0762 & 0.1881 & 0.1400 \\
DeepSeek-7B & 0.0000 & 0.0000 & 0.0000 \\
\bottomrule
\end{tabular}
\end{table}

Table~\ref{tab:small_llm_finetune_ablation} shows that LoRA fine-tuning does not consistently improve span-level causal extraction for the two strongest small LLMs. Mistral-7B performs better in its base setting, while Qwen-7B improves on Token Overlap F$_1$ after fine-tuning but does not improve on Exact Pair F$_1$. These results suggest that fine-tuning affects extraction behaviour unevenly.

\begin{table}[t]
\centering
\tiny
\setlength{\tabcolsep}{4pt}
\caption{Small-LLM causal cardinality ablation. $X_{\text{only}}$ contains single-pair causal spans, while $Y_{\text{only}}$ contains multi-pair causal spans. The combined row reports the best small-LLM result on the full extraction setting.}
\label{tab:small_llm_cardinality_ablation}
\begin{tabular}{llc}
\toprule
\textbf{Split} & \textbf{Best small-LLM config.} & \textbf{Pair F$_1$} \\
\midrule
single-pair & Mistral-7B CoT & 0.1044 \\
multiple-pair & Qwen-7B CoT & 0.1881 \\
Combined & Mistral-7B L2M & 0.1527 \\
\bottomrule
\end{tabular}
\end{table}

In Table~\ref{tab:small_llm_cardinality_ablation} we conduct a small-LLM causal cardinality ablation by disaggregating the extraction evaluation into single-pair causal spans, multi-pair causal spans and combined causal spans. This ablation isolates the effect of causal relation cardinality on extraction performance.

Table~\ref{tab:label_span_ablation} shows that label correctness is highest when extracted spans exactly match the gold spans. Causality classification drops more sharply under partial span matches than sententiality classification, suggesting that Explicit/Implicit labels are more sensitive to span-boundary errors.

\begin{table}[t]
\centering
\small
\resizebox{\columnwidth}{!}{%
\begin{tabular}{lrrrrrr}
\hline
\textbf{Metric} & $\textbf{C}_{\textbf{EM}}$ & $\textbf{C}_{\textbf{PM}}$ & $\textbf{C}_{\textbf{AM}}$ & $\textbf{S}_{\textbf{EM}}$ & $\textbf{S}_{\textbf{PM}}$ & $\textbf{S}_{\textbf{AM}}$ \\
\hline
Pairs       & 818    & 2,341  & 3,159  & 818    & 2,341  & 3,159  \\
Accuracy    & 79.22\% & 53.14\% & 59.89\% & 94.87\% & 76.80\% & 81.48\% \\
Macro F1    & 62.74\% & 53.10\% & 58.33\% & 50.95\% & 49.18\% & 50.45\% \\
Weighted F1 & 80.25\% & 53.30\% & 60.66\% & 96.79\% & 83.80\% & 87.33\% \\
\hline
\end{tabular}%
}
\caption{Model performance for causality and sententiality labeling across exact, partial, and all span-match subsets. $\textbf{C}$ = Causality, $\textbf{S}$ = Sententiality, $\textbf{EM}$ = Exact Match only, $\textbf{PM}$ = Partial Match only, and $\textbf{AM}$ = All Matches.}
\label{tab:label_span_ablation}
\end{table}

\end{document}